%% file: main.tex
\documentclass{article}
\usepackage{arxiv}

\usepackage[utf8]{inputenc} 
\usepackage[T1]{fontenc}    
\usepackage{url}            
\usepackage{booktabs}       
\usepackage{amsfonts}       
\usepackage{nicefrac}       
\usepackage{microtype}      
\usepackage{lipsum}
\usepackage{graphicx}

\usepackage[nocompress]{cite}
\usepackage[switch]{lineno}
\usepackage{times}
\usepackage{epsfig}
\usepackage{graphicx}
\usepackage{amsmath,amssymb,amsthm,amsfonts,fixmath}

\usepackage{algpseudocode,algorithm,algorithmicx}

\usepackage{array}
\newcolumntype{L}[1]{>{\raggedright\let\newline\\\arraybackslash\hspace{0pt}}m{#1}}
\newcolumntype{C}[1]{>{\centering\let\newline\\\arraybackslash\hspace{0pt}}m{#1}}
\newcolumntype{R}[1]{>{\raggedleft\let\newline\\\arraybackslash\hspace{0pt}}m{#1}}
  
\usepackage{diagbox}

\usepackage{stfloats}  

\usepackage{colortbl}

\newcommand{\dpar}[2]{\frac{\partial #1}{\partial #2}}

\renewcommand{\tan}[0]{\mbox{tan}}
\newcommand{\atan}[0]{\mbox{atan}}

\DeclareMathOperator*{\argmin}{argmin}

\newcommand{\setI}{\mathcal{I}}
\newcommand{\setX}{\mathcal{X}}

\newcommand{\realSet}{\mathbb{R}}

\theoremstyle{definition}

\newtheorem{proposition}{Proposition}

\renewcommand{\paragraph}[1]{{\vspace{2mm}\noindent\textbf{#1}}}

\usepackage{lipsum}
\usepackage{color}
\definecolor{mygray}{gray}{0.6}
\definecolor{gray}{RGB}{180,180,180}
\definecolor{lightgray}{RGB}{230,230,230}

\usepackage[pagebackref=true,breaklinks=true,colorlinks,bookmarks=false]{hyperref}

\title{\emph{Differential 3D Facial Recognition:} \\ Adding 3D to your State-of-the-Art 2D Method}
\author{
  J. Matias Di Martino \\
  Department of Physics \\
  Universidad de la Republica\\
  Montevideo, Uruguay \\
  \texttt{matiasdm@fing.edu.uy} \\
   \And
  Fernando Suzacq \\
  Department of Electrical Engineering\\
  Universidad de la Republica\\
  Montevideo, Uruguay \\
  fsuzacq@gmail.com \\
  \And
  Mauricio Delbracio\thanks{Now at Google Research.} \\
  Department of Electrical Engineering\\
  Universidad de la Republica\\
  Montevideo, Uruguay \\
  mdelbra@fing.edu.uy\\
  \And
  Qiang Qiu \\
  Department of Electrical Engineering\\
  Duke University \\
  Durham, NC, USA \\
  qiuqiang@gmail.com \\
  \And
  Guillermo Sapiro \\
  Department of Electrical Engineering\\
  Duke University \\
  Durham, NC, USA \\
  guillermo.sapiro@gmail.com \\
}

\begin{document}
\maketitle
\begin{abstract}
Active illumination is a prominent complement to enhance 2D face recognition and make it more robust, e.g., to spoofing attacks and low-light conditions. In the present work we show that it is possible to adopt active illumination to enhance state-of-the-art 2D face recognition approaches with 3D features, while bypassing the complicated task of 3D reconstruction. The key idea is to project over the test face a high spatial frequency pattern, which allows us to simultaneously recover real 3D information plus a standard 2D facial image. Therefore, state-of-the-art 2D face recognition solution can be transparently applied, while from the high frequency component of the input image, complementary 3D facial features are extracted. Experimental results on ND-2006 dataset show that the proposed ideas can significantly boost face recognition performance and dramatically improve the robustness to spoofing attacks.  
\end{abstract}

\section{Introduction}\label{sec:introduction}
\begin{figure}[htp]
\centering\includegraphics[width=.7\columnwidth]{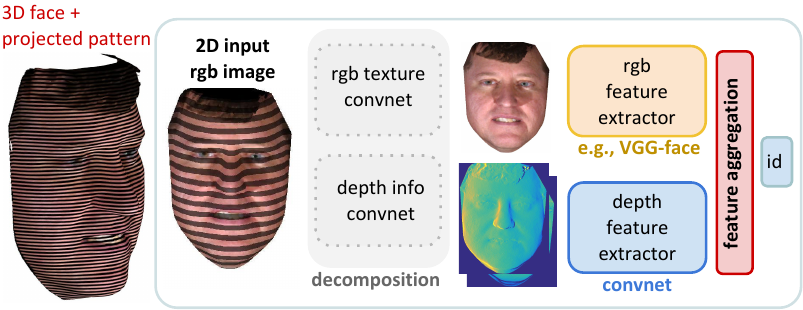}
\caption{Real 3D face recognition is possible by capturing one single RGB image if a high frequency pattern is projected. The low frequency components of the captured image can be fed into a state-of-the-art 2D face recognition method, while the high frequency components encode local depth information that can be used to extract 3D facial features. It is important to highlight that, in contrast with most existing 3D alternatives, the proposed approach provides real 3D information, not 3D hallucination from the RGB input. As a result, state-of-the-art 2D face recognition methods can be enhanced with real 3D information.} \label{fig:teaser}
\end{figure}

Two-dimensional face recognition has become extremely popular as it can be ubiquitously deployed and large datasets are available. 
In the past several years, tremendous progress has been achieved in making 2D approaches more robust and useful in real-world applications.
Though 2D face recognition has surpassed human performance in certain conditions, challenges remain to make it robust to facial poses, uncontrolled ambient illumination, aging, low-light conditions, and spoofing attacks \cite{Ding2016, kemelmacher2016megaface, Nech2017, taigman2014deepface}.
In the present work we address some of these issues by enhancing the captured RGB facial image with 3D information as illustrated in Figure~\ref{fig:teaser}.

High resolution cameras became ubiquitous, although for 2D face recognition, we only need a facial image of moderate or low resolution. 
For example latest phones frontal camera have a very high resolution (e.g., $3088\times2320$ pixels) while the resolution of the input to most face recognition systems is limited to $224\times224$ pixels \cite{arcface, parkhi2015deep, schroff2015facenet, taigman2014deepface, Zulqarnain2018}.
This means that, in the context of face recognition, we are drastically underutilizing most of the resolution of captured images. 
We propose an alternative to use the discarded portion of the spectra and extract real 3D information by projecting a high frequency light pattern.
Hence, a low resolution version of the RGB image remains approximately invariant allowing the use of standard 2D approaches, while 3D information is extracted efficiently from the local deformation of the projected patterns. 

The proposed solution to extract 3D facial features has key differences with the two common approaches presented in existing literature: 3D hallucination \cite{eigen2014depth, Huber2015, Liu2015, Pini2018} and 3D reconstruction \cite{Zafeiriou2013, Zou2005}. 
We will discuss these differences in detail in the following section.
We illustrate the main limitation of 3D hallucination in the context of face recognition in Figure~\ref{fig:3dillustration}, which emphasizes the lack of real 3D information on a standard RGB input image.
We demonstrate that it is possible to extract actual 3D facial features bypassing the ill-posed problem of explicit depth estimation.
Our contributions are summarized as follows:
\begin{itemize}
\itemsep 0.2em
	\item Analyzing the spectral content of thousands of facial images, we design a high frequency light pattern that simultaneously allow us to retrieve a standard 2D low resolution facial image plus a 3D gradient facial representation. 
	\item We propose an effective and modular solution that achieves 2D and 3D information decomposition and facial feature extraction in a data-driven fashion (bypassing a 3D facial reconstruction).
	\item We show that by defining an adequate distance function in the space of the feature embedding, we can leverage the advantages of both 2D and 3D features. We can transparently exploit existing state-of-the-art 2D methods and improve their robustness, e.g., to spoofing attacks. 
\end{itemize}	

\section{Related Work}
\label{sec:relatedWork}
To recognize or validate the identity of a subject from a 2D color photograph is a longstanding problem of computer vision and has been largely studied for over forty years~\cite{kaya1972basic, zhao2003face}. 
Recent advances in machine learning, and in particular, the success of deep neural networks, reshaped the field and yielded more efficient, accurate, and reliable 2D methods such as: ArcFace\cite{arcface}, VGG-Face~\cite{parkhi2015deep}, DeepFace~\cite{taigman2014deepface}, and FaceNet~\cite{schroff2015facenet}. 

In spite of this, spoofing attacks and variations in pose, expression and illumination are still active challenges and significant efforts are being made to address them~\cite{Cao2018, Hayat2017, He2018, Kumar2018, Lezama2017, Liu2018, Tran2017, Yu2017, Zhao2018, Zou2005}.
For example, Deng et al.~\cite{Deng2018} attempt to handle large pose discrepancy between samples. To that end, they propose an adversarial facial UV map completion GAN.  
Complementing previous approaches that seek for robust feature representations, several works propose more robust loss and metric functions \cite{Liu2017, Wang2018}. 

\paragraph{3D hallucination from single RGB.} To enhance 2D approaches a common trend is to hallucinate a 3D representation from an input RGB image which is used to extract 3D features \cite{blanz2003face, Dou2017, eigen2014depth, Huber2015, Liu2015, Pini2018}. 
For example, Cui et al.~\cite{Cui2018} introduce a cascade of networks that simultaneously recover depth from an RGB input while seeking for separability of individual subjects.
The estimated depth information is then used as a complementary modality to RGB. 
\begin{figure}
    \centering
    \includegraphics[width = .6\columnwidth]{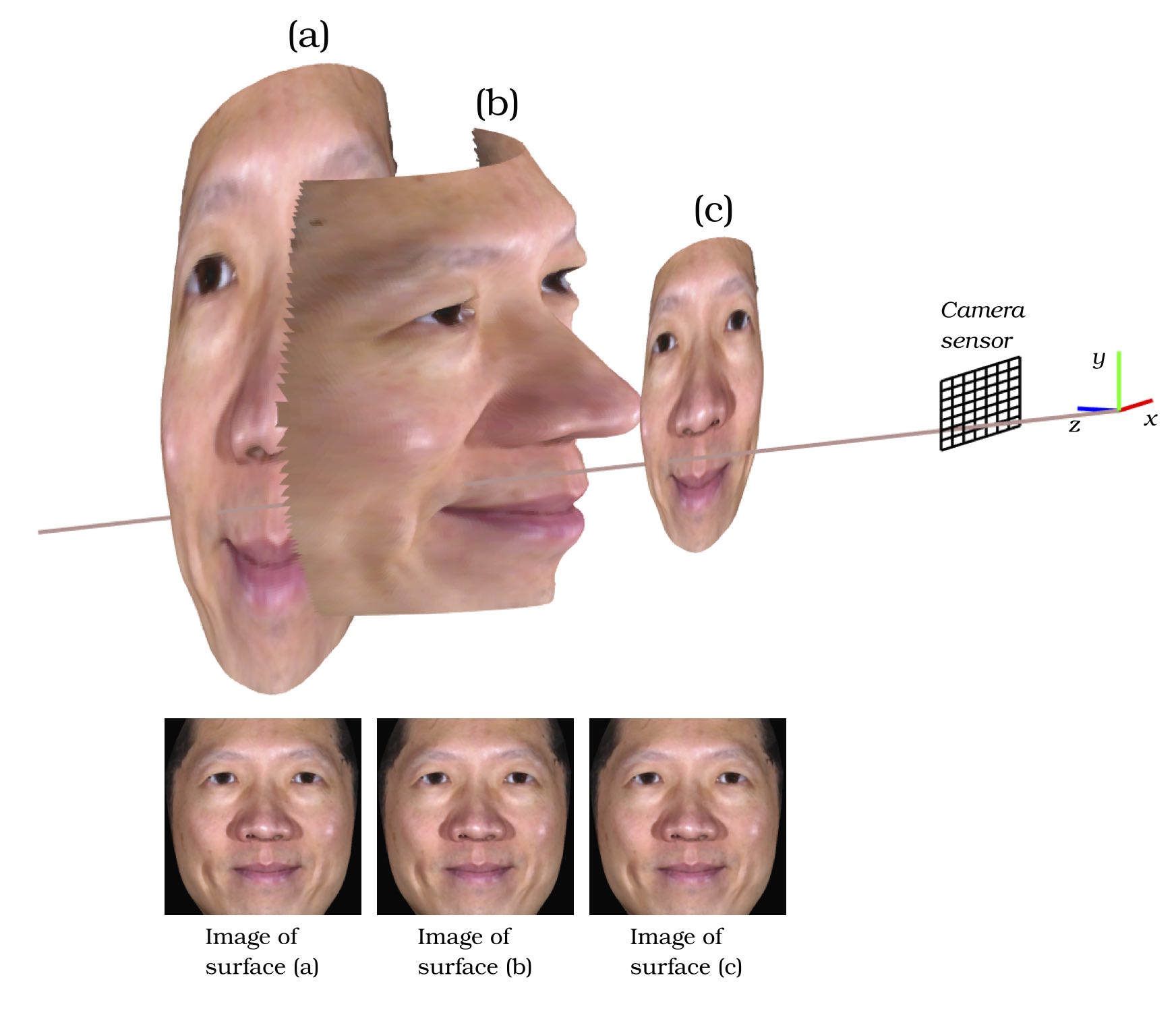}
    \caption{Illustration of three different 3D surfaces that look equivalent from a monocular view (single RGB image). On top, three surfaces (a), (b) and (c) are simulated, being (a) and (c) flat and (b) the 3D shape of a test subject. We use classic projective geometry \cite{hartley2003multiple} and simulate the image we obtain when photographing (a), (b) and (c) respectively. The resulting images are shown at the bottom. As we illustrate with this simple example, the relation between images and 3D scenes is not bijective and the problem of 3D hallucination is ill-posed. To overcome this, 3D hallucination solutions enforce important priors about the geometry of the scene. This is why we argue, that these methods do not really add to the face recognition task, actual 3D information. (A complementary example is presented in Figure~\ref{fig:3dmm} in the supplementary material).}
    \label{fig:3dillustration}
\end{figure}

\paragraph{3D face recognition.} 
The approaches described previously share an important practical advantage that at the same time is their weakness, they extract all the information from a standard (RGB) 2D photograph of the face. 
As depicted in~Figure~\ref{fig:3dillustration} a single image does not contain actual 3D information. 
To overcome this intrinsic limitation different ideas have been proposed and datasets with 3D facial information are becoming more popular~\cite{Zulqarnain2018}. 
For example, Zafeiriou et al.~\cite{Zafeiriou2013} propose a four-light source photometric stereo (PS). 
A similar idea is elaborated by Zou et al.~\cite{Zou2005} who propose to use active near-infrared illumination and combine a pair of input images to extract an illumination invariant face representation. 

Despite the previous mentioned techniques, performing a 3D facial reconstruction is still a challenging and complicated task. 
Many strategies have been proposed to tackle this problem, including time delay based~\cite{marks1992system}, image cue based \cite{eigen2015predicting, eigen2014depth,  laina2016deeper, prados2006shape, saxena2009make3d}, and triangulation based methods \cite{ayubi2010pulse, DiMartino2015one, li2014some, Rosman2016, zhang2006high}. 
Although there has been great recent development, available technology for 3D scanning is still too complicated to be ubiquitously deployed~\cite{di2018one, hartley2003multiple, zhang2010recent, zhang2013handbook}.

The proposed solution has two key features that make it, to the best of our knowledge, different from existing alternatives. (a) Because the projected pattern is of a high spatial frequency, we can recover a standard (low resolution) RGB facial image that can be fed into state-of-the-art 2D face recognition methods. (b) We avoid the complicated task of 3D facial reconstruction and instead, extract local 3D features from the local deformation of the projected pattern. In that sense our ideas can be implemented exploiting existing and future 2D solutions. In addition, our approach is different from those that hallucinate 3D information. As discussed before and illustrated in Figure~\ref{fig:3dillustration} this task requires a strong prior of the scene which is ineffective, for example, if a spoofing attack is presented (see the example provided in Figure \ref{fig:3dmm} in the supplementary material). 

\section{Proposed Approach}\label{sec:ProposedApproach}
\paragraph{Notation.} Let $\setI\subset\realSet^{H\times W\times C}$ denote the space of images with $H\times W$ pixels and $C$ color channels, and $\setX_{n}\subset\realSet^{n}$ a space of n-dimensional column vectors (in the context of this work associated to a facial feature embedding).
$\setI_{rgb}$ denotes the set of RGB images ($C=3$), while $\setI_{\nabla z}$ is used to denote the space of two channel images ($C=2$) associated to the gradient of a single-channel image $z\in\realSet^{H\times W \times 1}$. (The first/second channel represents the partial derivative with respect to the first/second coordinate.) 

\begin{figure*}[htp]
    \centering\includegraphics[width=.9\textwidth]{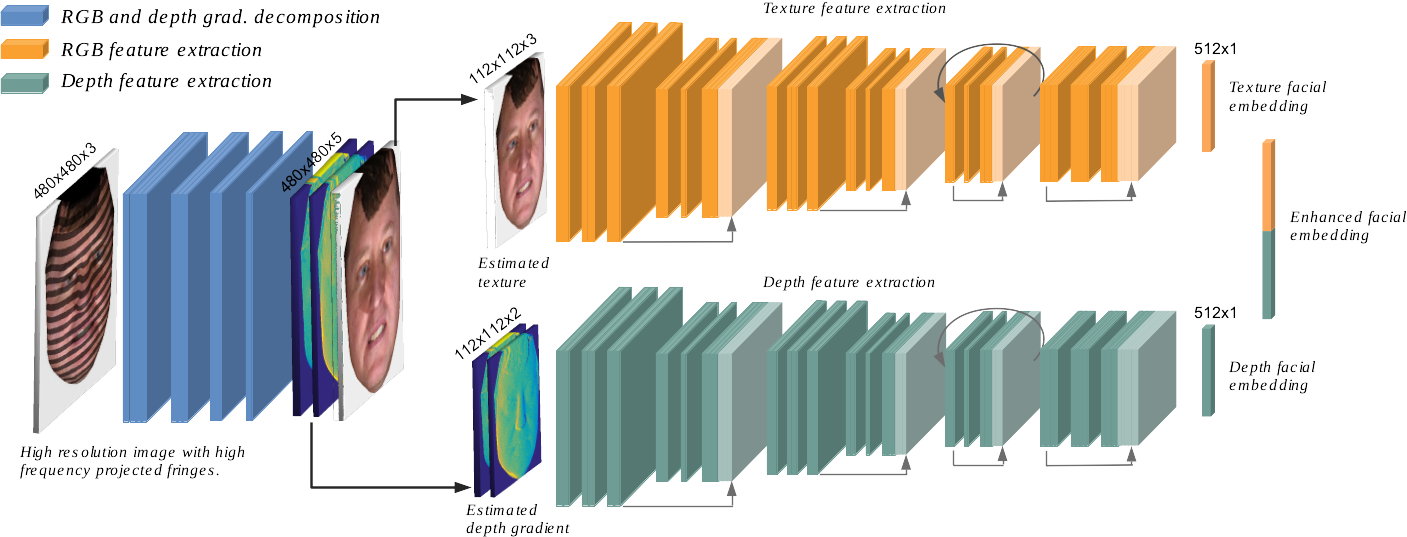}
    \caption{Architecture overview. First a network (illustrated in blue) is used to decompose the input image that contains overlapped high frequency fringes into a lower resolution (standard) texture facial image and depth gradient information. The former is used as the input of a state-of-the-art 2D face recognition DNN (yellow blocks). The depth information is fed to another network (green blocks) trained to extract discriminative (depth-based) facial features. Different network architectures are tested, we provide implementation details in Section~\ref{app:implementation_details} in the supplementary material.}\label{fig:proposedFramework}
\end{figure*}
\paragraph{Combining depth and RGB information.} The proposed approach consists of three main modules as illustrated in Figure~\ref{fig:proposedFramework}: $g:\setI_{rgb}\rightarrow \setI_{rgb}\times\setI_{\nabla z}$ performs a decomposition of the input image into texture and depth information, $f_{rgb}:\setI_{rgb}\rightarrow \setX_{n/2}$, and $f_{\nabla z}:\setI_{\nabla z}\rightarrow \setX_{n/2}$ extract facial features associated to the facial texture and depth respectively. 
These three components are illustrated in Figure~\ref{fig:proposedFramework} in blue, yellow, and green, respectively. 
(We decided to have three modules instead of a single \emph{end-to-end} design for several reasons that will be discussed below.) 

We denote the facial feature extraction from the input image as $f_{\theta} : \setI_{rgb} \rightarrow \setX_n$, where $f_\theta(I) = (f_{rgb}(I_{rgb}), f_{\nabla z}(I_{\nabla z}))^T$ with $\{I_{rgb},I_{\nabla z}\} = g(I)$. 
The subscript $\theta$ represent the parameters of the mapping $f$, which can be decomposed in three groups $\theta = (\theta_g, \theta_{rgb}, \theta_{\nabla z})$, associated to the image decomposition, RGB feature extraction, and depth feature extraction respectively. 
In the following we discuss how these parameters are optimized for each specific task, which is one of the advantages of formulating the problem in a modular fashion.  

Once texture and depth facial information is extracted into a suitable vector representation $x = f_{\theta}(I)$ (as illustrated in Algorithm~\ref{alg:facial_embedding}), we can select a distance measure $d:\setX_n\times\setX_n\rightarrow \realSet^+$ to compare facial samples and estimate whether they have a high likelihood of belonging to the same subject or not.
%
%
It is worth noticing that faces are embedded into a space in which the first half of the dimensions are associated to information extracted from the RGB representation while the other half codes depth information. 
These two sources of information may have associated different confidence levels (depending on the conditions at deployment). 
We address this in detail in Section~\ref{sec:distanceDesign} and propose an anisotropic distance adapted to our solution, and capable of leveraging the good performance of 2D solutions in certain conditions, while improving robustness and handling spoofing attacks in a continuous and unified fashion.
%
\begin{algorithm}
\begin{algorithmic}[1]
\Procedure{FacialEmbedding}{$I$}
\Statex Decompose the input image into texture and depth gradient information.
\State $\left\{ I_{rgb}, I_{\nabla z}\right\} = g(I)$
\Statex Extract facial information from each component.
\State $x_{rgb} = f_{rgb}(I_{rgb})$
\State $x_{\nabla z} = f_{\nabla z}\left(I_{\nabla z}\right)$
\Statex Combine texture and depth information.
\State $x = $ Concatenate$(x_{rgb}, x_{\nabla z})$
\State \textbf{return} $x$ \Comment{Facial embedding}
\EndProcedure
\end{algorithmic}
\caption{Compute 2D facial features enhanced with 3D information.}\label{alg:facial_embedding}
\end{algorithm}
%

\subsection{Pattern design.} \label{sec:pattern design}
When a pattern of light $p(x,y)$ is projected over a surface with a height map $z(x,y)$, it is perceived by a camera located along the $x$-axis with a deformation given by $p(x+\phi(x,y),y)$ ($\phi(x,y) \propto z(x,y)$). A detailed description of active stereo geometry is provided in the supplementary material Section~\ref{sec:reviewActiveStereo}. Let us denote $I_0(x,y)$ the image we would acquire under homogeneous illumination, and $p(x,y)$ the intensity profile of the projected light. Without loss of generality we assume the system baseline is parallel to the $x$ axis. The image acquired by the camera when the projected light is modulated with a profile $p(x,y)$ is
\begin{equation}\label{eq:pat_desing_1}
    I(x,y) = I_0(x,y)p(x+\phi(x,y), y).
\end{equation}

We will restrict to periodic modulation patterns and let $T$ denote the pattern spatial period, we also define $f_0 \stackrel{def}{=} \frac 1 T$. To simplify the system design and analysis, lets also restrict to periodic patterns that are invariant to the $y$ coordinate. In these conditions we can express $p(x,y) = \sum_{n=-\infty}^{+\infty} a_n\, e^{i2\pi n f_0 x}$ where $a_n$ represent the coefficients of the Fourier series of $p$. (Note that because of the invariance with respect to the $y$ coordinate, the coefficients $a_n$ are constant instead of a function of $y$.) Equation~\eqref{eq:pat_desing_1} can be expressed as 
\begin{equation}\label{eq:pat_design_2}
    I(x,y) = \sum_{n=-\infty}^{+\infty} I_0(x,y)\, a_n\, e^{i2\pi n f_0 (x+\phi(x,y))}.
\end{equation}
Defining $q_n(x,y)\stackrel{def}{=}I_0(x,y)\,a_n\,e^{i2\pi n f_0 \phi(x,y)}$, Equation~\eqref{eq:pat_design_2} can be expressed as \cite{takeda1983fourier}
\begin{equation}\label{eq:pat_design_3}
    I(x,y) = \sum_{n=-\infty}^{+\infty} q_n(x,y) e^{i2\pi n f_0 x}.
\end{equation}
Applying the 2D Fourier Transform (FT) in both sides of Equation~\eqref{eq:pat_design_3} and using standard properties of the FT \cite{distributions} we obtain
\begin{equation}\label{eq:pat_design_4}
    \tilde{I}(f_x,f_y) = \sum_{n=-\infty}^{+\infty} \tilde{q_n}(f_x - nf_0, f_y).
\end{equation}
We denote as $\tilde{I}$ the FT of $I$ and use $(f_x,f_y)$ to represent the 2D frequency domain associated to $x$ and $y$ axis respectively. 
\begin{figure}[htp]
\centering\includegraphics[width = .6\columnwidth]{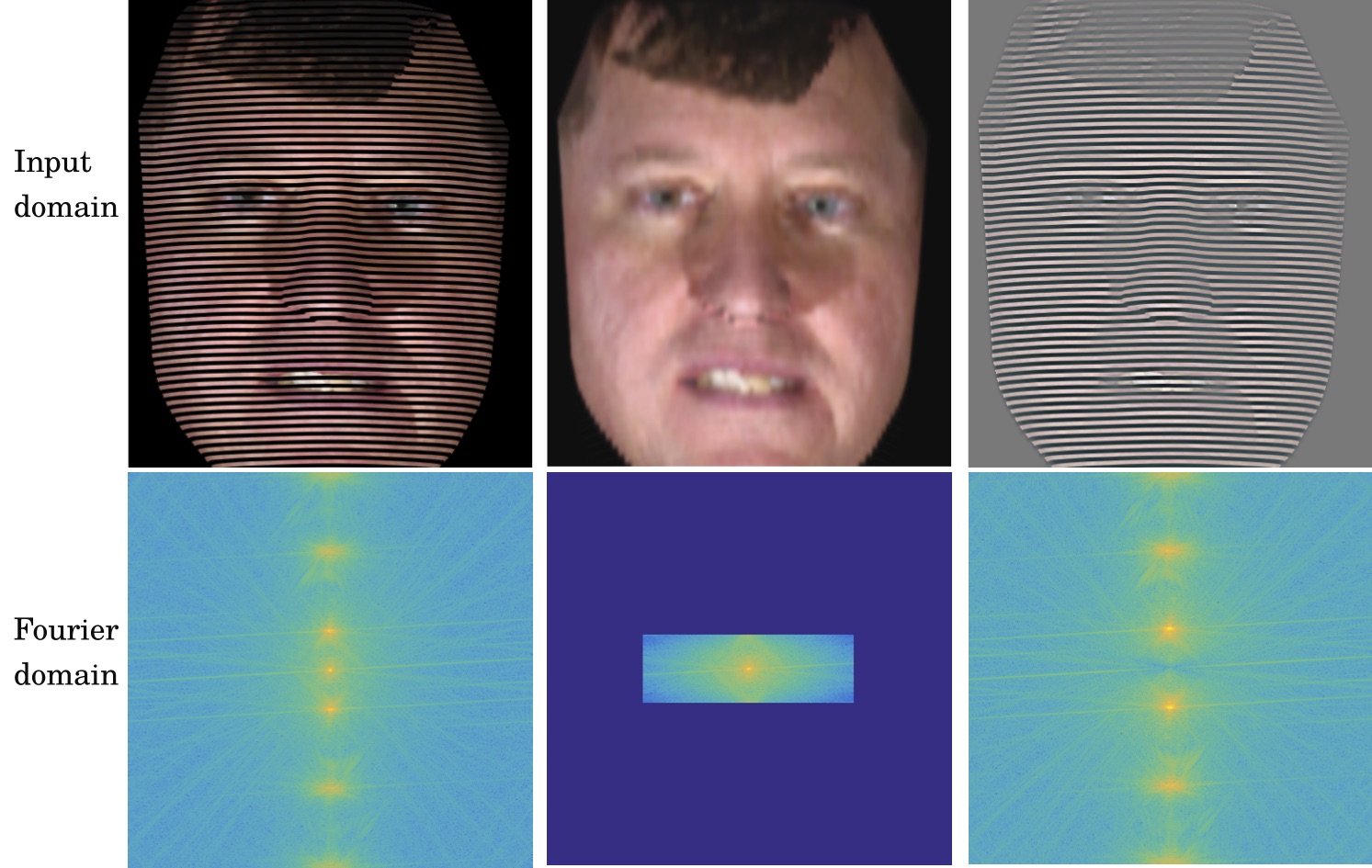}
\caption{2D plus real 3D in a single rgb image. The first column illustrates the RGB image acquired by a (standard) camera when horizontal stripes are projected over the face. The second column isolates the low frequency components of the input image, and the third column corresponds to the residual high frequency components. (In all the cases the absolute value of the Fourier Transform is represented in logarithmic scale). As can be seen, high frequency patterns can be used to extract 3D information of the face (third column) while preserving a lower resolution version of the facial texture (middle column).} \label{fig:fringesDescomposition}
\end{figure}

Equation~\eqref{eq:pat_design_4} shows that the FT of the acquired image can be decomposed into the components $\tilde{q_n}$ centered at $(nf_0,0)$. In the context of this section, we refer to a function $h(x,y)$ being smooth if
\begin{equation}
    \frac{\|\tilde{h}(f_x,f_y)\|}{\|\tilde{h}(0,0)\|}<10^{-3}\ \ \  \forall \ \ \ |f_x|>\frac{f_0}{2} .
\end{equation}
Assuming $I_0(x,y)$ and $\phi(x,y)$ are smooth (we empirically validate this hypothesis below), the components $\tilde{q_n}$ can be isolated as illustrated in Figure~\ref{fig:fringesDescomposition}. 
The central component is of particular interest, $q_0(x,y) = a_0\, I_0(x,y)$ captures the facial texture information and can be recovered from $I(x,y)$ if $f_0$ is large enough (we provide a more precise quantitative analysis in what follows).
On the other hand, relative (gradient) 3D information can be retrieved from the components $\{q_0, q_1\}$ as we show in Proposition~\ref{prop:gradient_depth}.
\begin{proposition}\label{prop:gradient_depth}
Gradient depth information is encoded in the components $\{q_0(x,y), q_1(x,y)\}$.
\end{proposition}
\begin{proof}
We define the wrapping function $\mathcal{W}(u)=\atan(\tan(u))$. This function wraps the real set into the interval $\left(-\pi/2,\pi/2\right]$ \cite{ghiglia1998unwrapping}. This definition can be extended to vector inputs wrapping the modulus of the vector field while keeping its direction unchanged, i.e., $\mathcal{W}(\vec{u})=\frac{\mathcal{W}(\|\vec{u}\|)}{\|\vec{u}\|}\vec{u}$ if $\|\vec{u}\|\neq 0$ and $\mathcal{W}(\vec{u})=\vec{0}$ if $\|\vec{u}\| = 0$. 
From $q_1(x,y)$ and $q_0(x,y)$ we can compute\footnote{We assume images are extended in an even fashion outside the image domain, to guaranteed that $a_1\in\realSet$ and avoid an additional offset term.} 
\begin{equation}\label{eq:prop_prof_1}
    \phi_{\mathcal{W}}(x,y) = \frac{1}{2\pi f_0}\, \atan\left(\frac{\mbox{Im}\{\frac{q_1(x,y)}{q_0(x,y)}\}}{\mbox{Re}\{\frac{q_1(x,y)}{q_0(x,y)}\}}\right)
\end{equation}
where $\phi_{\mathcal{W}}$ denotes the wrapped version of $\phi$. Moreover, $\phi_{\mathcal{W}}(x,y)=\phi(x,y) + \pi k(x,y)$ with $k(x,y)\in\mathbb{N}$ (wrapping introduces shifts of magnitude multiple of $\pi$). Computing the gradient both sides leads to
$\nabla\phi_{\mathcal{W}}(x,y)=\nabla\phi(x,y) + \pi \nabla k(x,y)$ where $\|\nabla k(x,y)\|\in\mathbb{N}$. Assuming the magnitude of the gradient of $\phi(x,y)$ is bounded by $\pi/2$ and considering that $\|\nabla k(x,y)\|\in\mathbb{N}$, we can apply the wrapping function both sides of the previous equality to obtain $\mathcal{W}(\nabla\phi_{\mathcal{W}})(x,y)=\nabla\phi(x,y)$ which proves (recall Equation~\eqref{eq:prop_prof_1}) that the gradient of $\phi$ can be extracted from the components $q_0$ and $q_1$. To conclude the proof, we use the property of linearity of the gradient operation and the fact that $\phi(x,y)$ is proportional to the depth map of the scene (see Equation~\eqref{eq:distarityvsdepth} and Section~\ref{sec:reviewActiveStereo} in the supplementary material). 
\end{proof}
\begin{figure}
    \centering
    \includegraphics[width=.6\columnwidth]{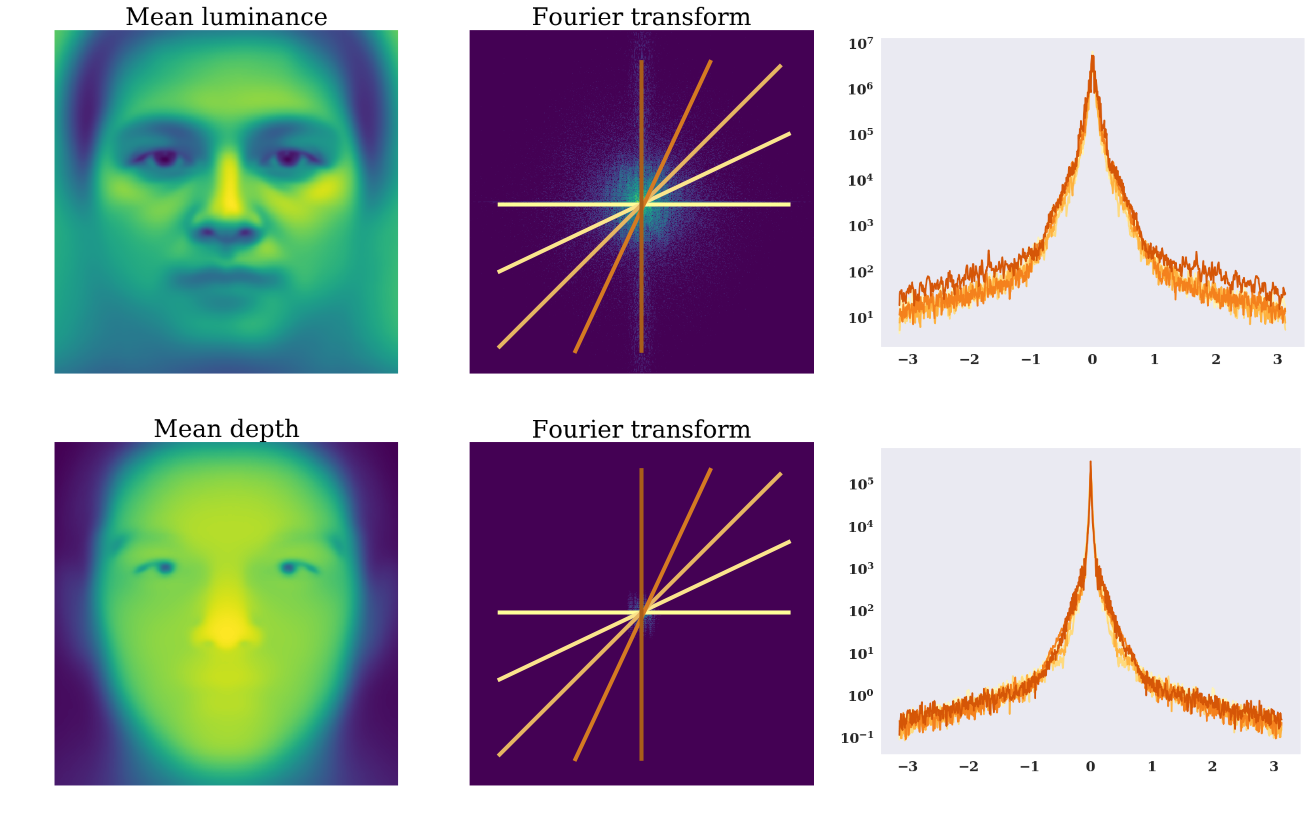}
    \caption{Faces average spectral content. The first column illustrates the mean luminance and depth map for the faces in the  dataset ND-2006. The second column shows the mean Fourier Transform of the faces luminance and depth respectively. The third column shows the profile across different sections of the 2D Fourier domain. Columns two and three represent the absolute value of the Fourier transform in logarithmic scale. Faces are registered using the eyes landmarks and the size normalized to $480\times480$ pixels.}
    \label{fig:face_spectrum}
\end{figure}

\paragraph{Analytic versus data-driven texture and gradient depth extraction} The previous analysis shows that closed forms can be obtained to extract texture and depth gradient information. However, to compute these expressions is necessary to isolate different spectral components $\tilde{q_n}$. To that end, filters need to be carefully designed. The design of these filters is challenging, e.g., one need to control over-smoothing versus introducing ringing artifact which are drastically amplified by a posterior gradient computation \cite{DiMartino2015one, zhang2006high}. To overcome these challenges, we chose to perform a depth (gradient) and texture decomposition in a data-driven fashion, which as we show in Section~\ref{sec:Experiments}, provides an efficient and effective solution.  

\paragraph{Bounds on $f_0$ and optimal spectral orientation.} As discussed above, the projected pattern $p(x,y)$ should have a large fundamental frequency $f_0$. In addition, the orientation of the fringes and the system baseline can be optimized if faces present a narrower spectral content in a particular direction. We study the texture and depth spectrum of the facial images of ND-2006 dataset (this dataset provides ground truth facial texture and depth information). 
We observed (see Figure~\ref{fig:face_spectrum}) that for facial images sampled at a $480\times480$ spatial resolution, most of the energy is concentrated in a third of the discrete spectral domain (observe the extracted one dimensional profiles of the spectrum shown at the left side of Figure~\ref{fig:face_spectrum}). 
In addition, we observe that the spectral content of facial images is approximately isotropic. 
See, for example, Figure~\ref{fig:face_spectrum} and observe how for 1-dimensional sections across different orientations the 2D spectra envelope is almost constant. 
We conclude that the orientation of the fringes does not play a significant role in the context of facial analysis. 
In addition, we conclude that the fringes width should be smaller than $7$mm (distance measure over the face).\footnote{This numerical results is obtained by approximating the bounding box of the face as a $20cm\times 20cm$ region, sampled with $480\times 480$ pixels which corresponds to a pixel length of $2.4mm$, a third of the spectral band correspond to signal of a period of $6$ pixels which leads to a binary fringe of at least $7.2mm$ wide.} 

\subsection{Network training and the advantages of modularity.} \label{sec:networkDesign}
As described previously, the parameters of the proposed solution can be split in three groups $\theta = (\theta_g, \theta_{rgb}, \theta_{\nabla z})$. 
This is an important practical property and we designed the proposed solution to meet this condition (in contrast to an end-to-end approach).

Let us define $\mathcal{B}_{1}$, $\mathcal{B}_{2}$, and $\mathcal{B}_{3}$ three datasets containing ground truth depth information, ground truth identity for rgb facial images, and ground truth identity for depth facial images, respectively. 
More precisely, $\mathcal{B}_{1}=\{(I_i(x,y), {I_0}_i(x,y), z_i(x,y)),\ i=1,\,...,\,n_1\}$, $\mathcal{B}_{2}=\{({I_0}_i(x,y), y_i),\ i=1,\,...,\,n_2\}$, and $\mathcal{B}_{3}=\{(z_i(x,y), y_i),\ i=1,\,...,\,n_3\}$, where $I_i(x,y)$ denotes a (facial or generic) RGB image acquired under the projection of the designed pattern, ${I_0}_i(x,y)$ represents (facial or generic) standard RGB images, $z_i(x,y)$ denotes a gray image representing the depth of the scene, and $y_i$ a scalar integer representing the subject id. 

We denote as $\{g_1(I), g_2(I)\} = g(I)$ the RGB and gradient depth components estimated by the decomposition operation $g$. We partitioned the parameters of $g$ into two sets of dedicated kernels $\theta_g = \{{\theta_g}_1, {\theta_g}_2\}$, the first group focuses on retrieving the texture component while the second group retrieves the depth gradient. These parameters can be optimized as 
\begin{eqnarray}\label{eq:loss_g}
    {\theta_g}_1 = \argmin \sum_{({I_0}_i,I_i)\in\mathcal{B}_1} \|g_1(I_i) - {I_0}_i \|_2^2 \\
    {\theta_g}_2 = \argmin \sum_{({z}_i,I_i)\in\mathcal{B}_1} \|g_2(I_i) - \nabla z_i \|_2^2.
\end{eqnarray}
(We also evaluated training a shared set of kernels trained with an unified loss, this alternative is harder to train in practice, due to the natural difference between the dynamic range and sparsity of gradient images compared with texture images.) 

For texture and depth facial feature extraction, we tested models inspired in the Xception architecture \cite{chollet2017xception}). Additional details are provided in the supplementary material Section \ref{app:implementation_details}. 
To train these models we add an auxiliary fully connected layer on top of the facial embedding (with as many neurons as identities in the train set) and minimize the cross-entropy between the ground truth and the predicted labels. More precisely, let us denote $\hat{f}_{rgb}(I_{rgb})=[p_1,...,p_c]$ the output of the fully connected layer associated to the embedding $f_{rgb}(I_{rgb})$ where $p_i$ denotes the probability associated to the id $i$, 
\begin{eqnarray}
    \theta_{rgb} = \argmin \sum_{({I_0}_i, y_i)\in\mathcal{B}_2} \sum_{c} -{\bf{1}}_{y_i=c} \log(\hat{f}_{rgb}({I_0}_i)[c]) \\
    \theta_{\nabla z} = \argmin \sum_{({z}_i, y_i)\in\mathcal{B}_3} \sum_{c} -{\bf{1}}_{y_i=c} \log(\hat{f}_{\nabla z}(\nabla z_i)[c])
\end{eqnarray}
where ${\bf{1}}_{y_i=c}$ denotes the indicator function. (Of course one can choose other alternative losses to train these modules, see e.g., \cite{arcface,Liu2017, Wang2018, Zheng2018}.)

As described above, the proposed design allows to leverage information from three types of datasets ($\mathcal{B}_1$, $\mathcal{B}_2$, $\mathcal{B}_3$). This has an important practical advantage as 2D facial and 3D generic datasets are more abundant, and the pattern dependant set $\mathcal{B}_1$ can be of modest size as $\#(\theta_g)\ll\#(\theta_{rgb})$.

\subsection{Distance design.} \label{sec:distanceDesign}
Once different modules are set we can compute the facial embedding of test images following the procedure described in Algorithm~\ref{alg:facial_embedding}. Let us define $x^a\in\setX_n$ and $x^b\in\setX_n$ the feature embedding of two facial images $I_a$ and $I_b$ respectively. Recall that the first $n/2$ elements of $x$ are associated to features extracted from (a recovered) RGB facial image while the remaining elements are associated to depth information, i.e., $x = (x_{rgb}[1],..., x_{rgb}[n/2], x_{\nabla z}[1], ..., x_{\nabla z}[n/2])^T$. 

We define the distance between two feature representations $x^a = (x_{rgb}^a, x_{\nabla z}^a)$, and $x^b = (x_{rgb}^b, x_{\nabla z}^b)$ as
\begin{equation}\label{eq:distance}
\begin{array}{l}
    \displaystyle d_{\alpha,\beta,\gamma}(x^a,x^b) \stackrel{def}{=}  (1-\gamma)\, d_c(x^a_{rgb},\,x^b_{rgb}) \vspace{2mm} \\
     \ \ +\ \gamma\, d_c(x^a_{\nabla z},\,x^b_{\nabla z}) \left(1 + \left(\frac{d_c(x^a_{\nabla z},\,x^b_{\nabla z})}{\beta}\right)^\alpha \right).
\end{array}
\end{equation}
$d_c:\setX_{n/2}\times\setX_{n/2}\rightarrow[0, 1]$ denotes the cosine distance, $\gamma\in[0,1]$ sets the relative weight of RGB and depth features, and $\alpha,\beta \in \realSet$ define a non-linear response for the distance  between depth features. As we will describe in the following, this provides robustness against common cases of spoofing attacks.

Intuitively, $\gamma$ allows us to set the relative confidence associated to RGB and depth features, for example, $\gamma=1/2$ gives the same weight to RGB and depth features, while $\gamma=0$ ($\gamma=1$) ignores the distance between samples in the depth (RGB) embedding space. This is important in practice, as is common to obtain substantially more data to train RGB models than depth ones ($|\mathcal{B}_2| \gg |\mathcal{B}_3|$). This suggests that in good test conditions (e.g., good lighting) one may trust more RGB features over depth features ($\gamma<1/2$). As we will empirically show in the following section, when two facial candidates are compared, $d_{\alpha,\infty,\gamma}(x^a,x^b) =  (1-\gamma)\, \delta(x^a_{rgb},\,x^b_{rgb}) + \gamma\, \delta(x^a_{\nabla z},\,x^b_{\nabla z})$ is an effective distance choice. However, it does not handle robustly common cases of spoofing attacks. The most common deployments of spoofing attacks imitate the facial texture more accurately than the facial depth \cite{OULU_NPU_2017, Liu2018spoofing, Zhang2016}, therefore, the global distance between two samples should be large when the distance of the depth features is large (i.e. above a certain threshold). To that end, we introduce an additional non-linear term controlled by parameters $\beta$ and $\alpha$, for $\delta(x^a_{\nabla z}, x^b_{\nabla z}) < \beta$ the standard cosine distance dominates while for large values the distance will be amplified in a non-linear fashion. 

\section{Experiments and Discussion}\label{sec:Experiments}
%
\paragraph{Data.} Three public dataset are used for experimental validation: FaceScrub \cite{FaceScrub}, CASIA Anti-spoofing \cite{zhang2012face}, and ND-2006 \cite{nd2006}. FaceScrub contains $100k$ RGB (2D) facial images of $530$ different subjects, and is used to train the texture-based facial embedding. CASIA dataset contains $150$ genuine videos (recording a person) and $450$ videos of different types of spoofing attacks, the data was collected for $50$ subjects. We use this dataset to simulate and imitate the texture properties of images of spoofing attacks. ND-2006 is one of the larges publicly available datasets with 2D and 3D facial information, it contains $13k$ images of $888$ subjects. We used this set to demonstrate that differential 3D features can be extracted from a single RGB input, to compare RGB features with 3D features extracted from the differential 3D input, and to show that when 2D and 3D information is properly combined, the best properties of each can be obtained.  

\paragraph{Texture and differential 3D decomposition.}
In Section~\ref{sec:pattern design} we discussed how real 3D information and texture information can be coded and later extracted using a single RGB image. In addition, we argue that this decomposition can be learned efficiently and effectively in a data-driven fashion.
To that end, we tested simple network architectures composed of standard convolutional layers (a full description of these architectures and the training protocols are provided as supplementary material). Using ground truth texture and depth facial information, we simulated the projection of the designed pattern over the $888$ subjects provided in ND-2006 dataset. Illustrative results are presented in Figure~\ref{fig:nd2006patterns} and in the supplementary material. The 3D geometrical model and a detailed description of the simulation process is provided in Section~\ref{app:sec:sim_proj_pat}. Though the simulation of the deformation of a projected pattern can be computed in a relatively simple manner (if the depth information is known), the inverse problem is analytically hard \cite{DiMartino2015one, Rosman2016, zhang2013handbook}. 
\begin{figure}
    \centering
    \includegraphics[width=.6\columnwidth]{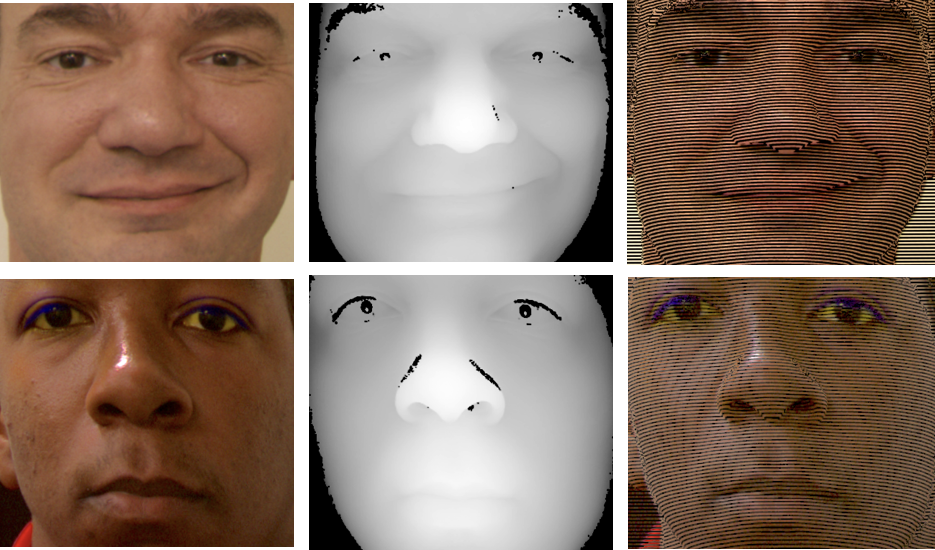}
    \caption{Active light projection. From left to right: ground truth RGB facial image, 3D facial scanner, and finally the image we would acquire if the designed high frequency pattern is projected over the face. Two random samples from ND-2006 are illustrated.} 
    \label{fig:nd2006patterns}
\end{figure}

\begin{figure}[htp]
\centering\includegraphics[width=.6\columnwidth]{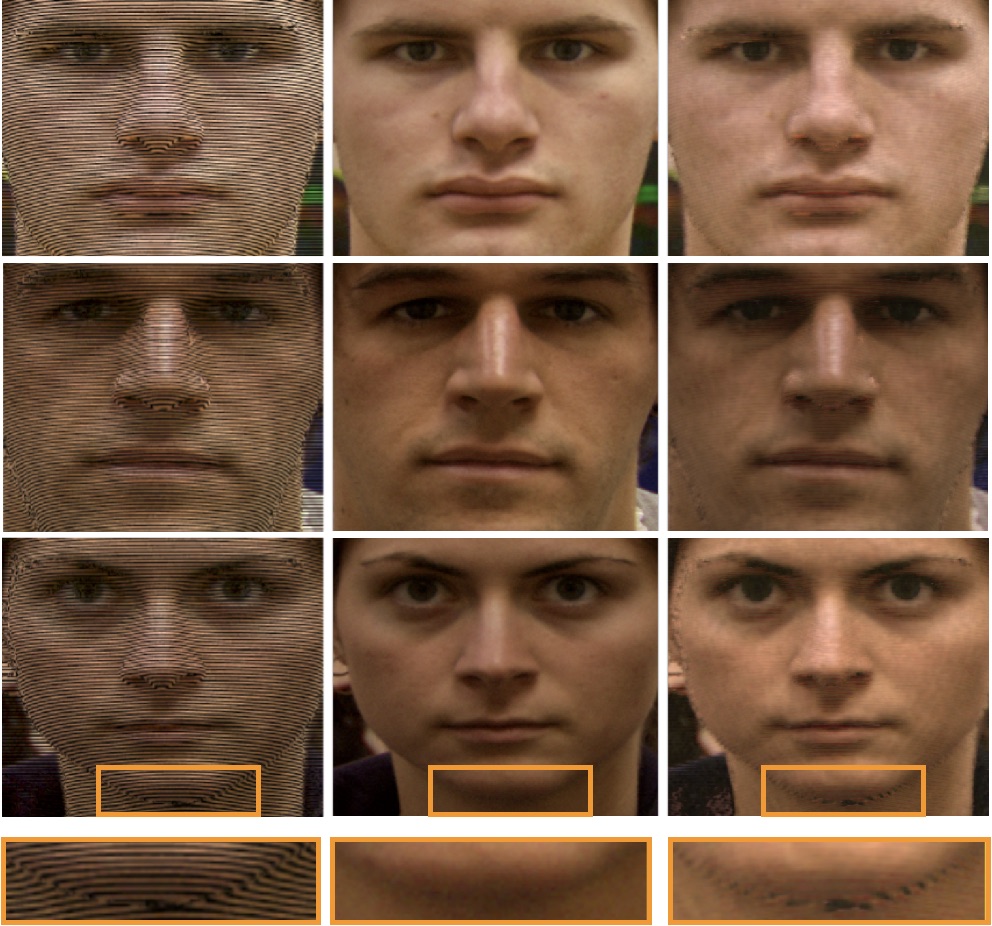}
\caption{Examples of the facial texture recovered from the image with the projected pattern. The first column, shows the input image (denoted as $I$ in Algorithm~\ref{alg:facial_embedding}). The second column shows the ground truth, and the third column the texture recovered by the network $I_{rgb}$. This examples are from the test set and the images associated to these subjects were never seen during the training phase. }\label{fig:texture_descomposition}
\end{figure}

Despite the previous, we observed that a stack of convolutional layers can efficiently learn how to infer from the image with the projected pattern, both depth gradient information, and the standard (2D) facial image. Figure~\ref{fig:texture_descomposition} illustrates some results for subjects in the test set. The first column corresponds to the input to the network, the second column the ground truth texture information, and the third column the retrieved texture information. The architecture of the network and the training protocol is described in detail in the supplementary material Section~\ref{app:implementation_details}. As we can see in the examples illustrated in Figure~\ref{fig:texture_descomposition}, an accurate low resolution texture representation of the face can be achieved in general, and visible artifact are observed only in the regions where the depth is discontinuous (see for example, the regions illustrated at the bottom of Figure~\ref{fig:texture_descomposition}). 

\begin{figure}[htp]
\hspace{-2mm}\includegraphics[width=\columnwidth]{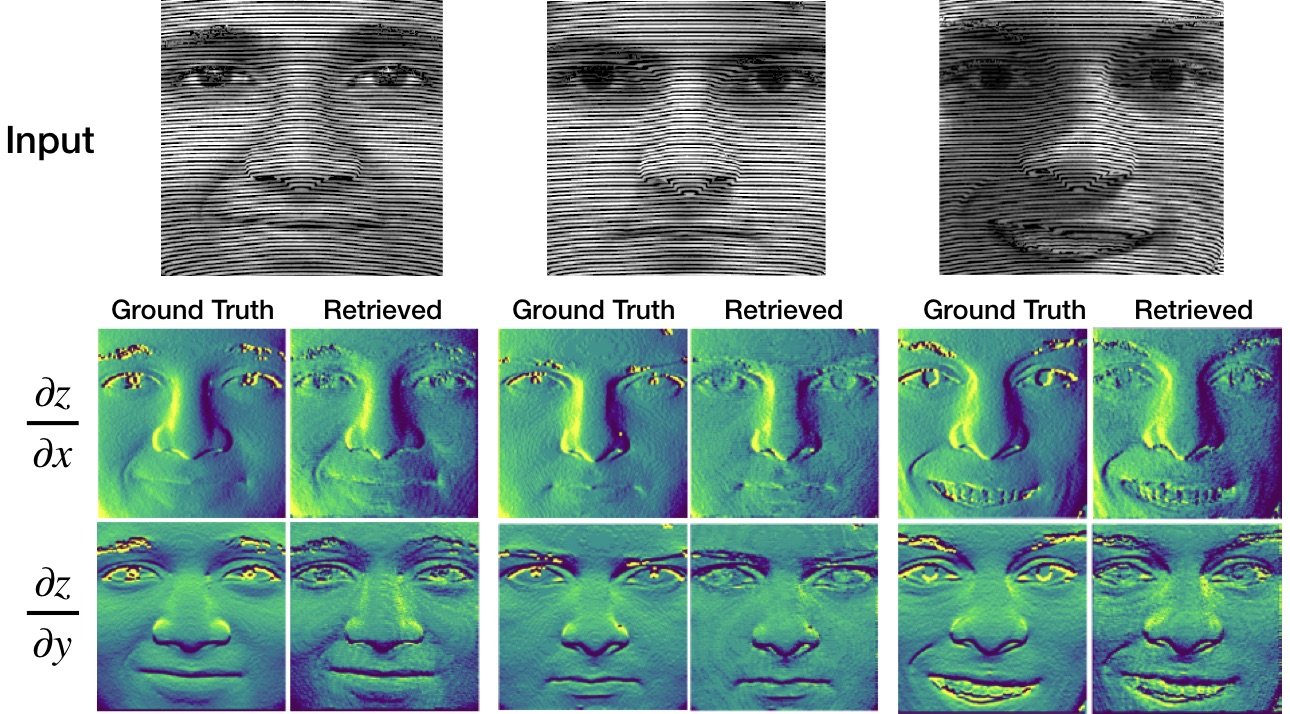}
\caption{Differential depth information extracted from the image with the projected pattern. The first row illustrates the input image (depth information can be extracted from a gray version of the input as the designed patter is achromatic). The second and third row show the ground truth and the retrieved $x$ and $y$ partial derivatives of the depth respectively.} \label{fig:depth_gradient_descomposition}
\end{figure}

Figure~\ref{fig:depth_gradient_descomposition} illustrates the ground truth and the retrieved depth gradient (again, for random samples from the test set). To estimate the 3D information, we feed to a different branch of convolutional layers the gray version of the input image. These layers are fully described in the supplementary material Table~\ref{tab:DescompositionNetwork}. A gray input image is considered instead of a color one because the projected pattern is achromatic, and therefore, no 3D information is encoded in the colors of the image. In addition, we crop the input image to exclude the edges of the face. (Facial registration and cropping is performed automatically using dlib \cite{king2009dlib} facial landmarks.) As discussed in Section~\ref{sec:ProposedApproach}, and in particular, in the proof of Proposition~\ref{prop:gradient_depth}, the deformation of the projected fringes only provide local gradient information if the norm of the gradient of the depth is bounded. In other words, where the scene present depth discontinuities, no local depth information can be extracted by our proposed approach. This is one of the main reasons why differential 3D information can be exploited for face recognition, while bypassing the more complicated task of a 3D facial reconstruction. 

\begin{figure}[htp]
\centering\includegraphics[width=.7\columnwidth]{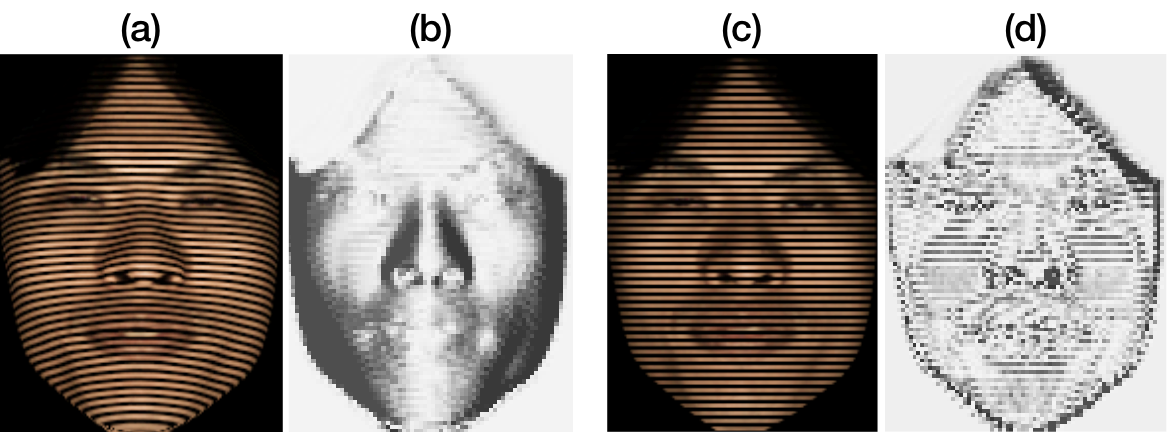}
\caption{Is the network really extracting depth information? In this figure we show the output of the network for two inputs generated using identical facial texture but different depth ground truth data. (a) Image obtained when the projected pattern is projected over the face with the real texture and the real 3D profile. (b) Output of the network when we input (a) (only the x-partial derivative is displayed for compactness). (c) Image obtained when the projected pattern is projected over a flat surface with the texture of the real face. (d) Output of the network when the input is (c). None of these images were seen during training.}\label{fig:depth_extraction_verification}
\end{figure}

One of the advantages of the proposed approach is that it extracts local depth information, and therefore, the existence of depth discontinuities does not affect the estimation on the smooth portion of the face. This is illustrated in Figure~\ref{fig:depth_extraction_verification} (a)-(b), where a larger facial patch is fed into the network. The decomposition module is composed exclusively of convolutional layers, and therefore, images of arbitrary size can be evaluated. Figure~\ref{fig:depth_extraction_verification}-(a) shows the input to the network, and Figure~\ref{fig:depth_extraction_verification}-(b) the first channel of the output (for compactness we display only the x-partial derivative). As we can see, the existence of depth discontinuities does not affect the prediction in the interior of the face (we consider the prediction outside this region as noise and we replace it by $0$ for visualization). 

Several algorithms have been proposed to hallucinate 3D information from a 2D facial image \cite{eigen2014depth, Huber2015, Liu2015, Pini2018}. In order to verify that our decomposition network is extracting real depth information (in lieu of hallucinating it from texture cues), we simulated an image where the pattern is projected over a surface with identical texture but with a planar 3D shape (as in the example illustrated in Figure~\ref{fig:3dillustration}). Figure~\ref{fig:depth_extraction_verification} (a) shows the image acquired when the fringes are projected over the ground truth facial depth, and (c) when instead the depth is set to $0$ (without modifying the texture information). The first component of the output (x-partial derivative) is shown in (b) and (d), as we can see, the network is actually extracting true depth information (from the deformation of the fringes) and not hallucinating 3D information from texture cues. (As we will see next, this property is particularly useful for joint face recognition and spoofing prevention.) 

\begin{figure*}
    \centering
    \includegraphics[width=\textwidth]{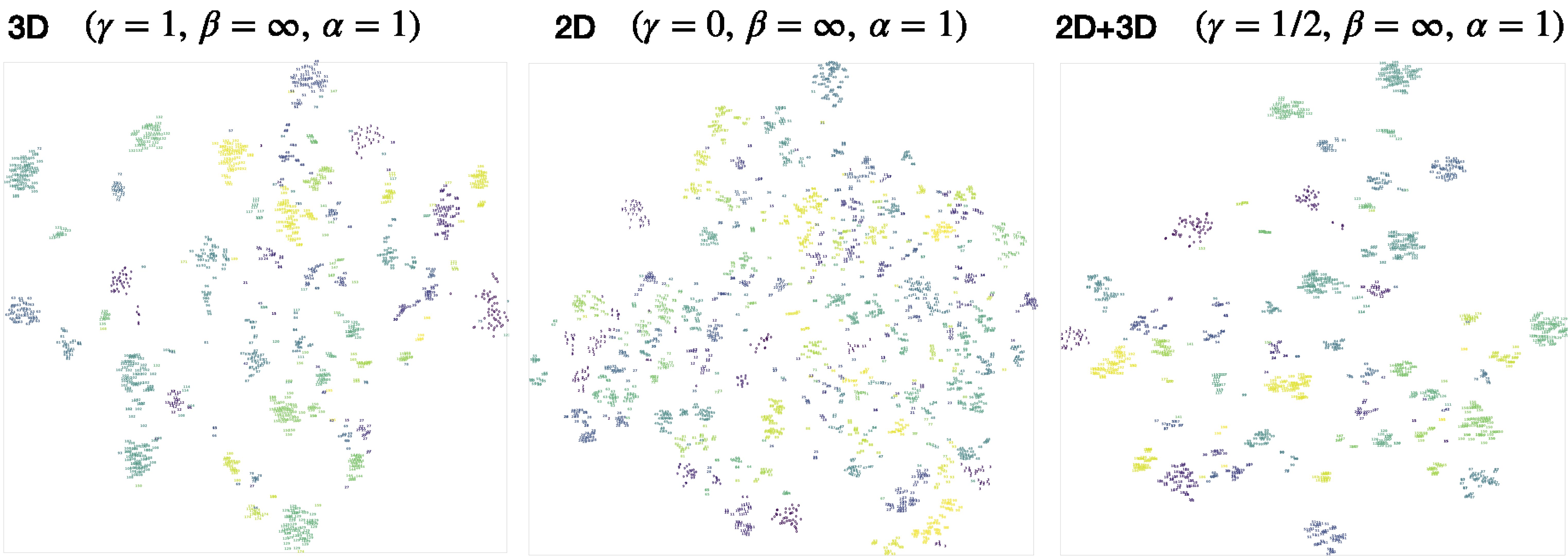}
    \caption{Facial features low dimensional embedding (for visualization purposes only). We illustrate texture-based and depth-based features in a low dimensional embedding space. A random set of subject of the test set is shown. From left to right: the embedding of depth-features, texture-based features, and finally, the combination of texture and depth features. t-SNE \cite{maaten2008visualizing} algorithm is used for the low-dimensional embedding.}\label{fig:embedding}
\end{figure*}

\paragraph{2D and 3D face recognition.} 
Once the input image is decomposed into a (standard) texture image and depth gradient information, we can proceed to extract 2D and 3D facial features from each component. To this end, state-of-the-art network architectures are evaluated. Our method is agnostic to the RGB and depth feature extractors, moreover, as the retrieved texture image is close to a standard RGB facial images (in sense of the L2-norm), any pre-train 2D feature extractor can be used (e.g., \cite{arcface, parkhi2015deep, schroff2015facenet, taigman2014deepface, Zulqarnain2018}). In the experiments presented in this section we tested a network based on the Xception architecture \cite{chollet2017xception} (details are provided as supplementary material). For the extraction of texture features, the network is trained using FaceScrub \cite{FaceScrub} dataset (as we previously described, this is a public dataset of 2D facial images). The module that extracts 3D facial features is trained using $2/3$ of the subjects of ND-2006 dataset, leaving the remaining subjects exclusively for testing. The output of each module is a 512-dimensional feature vector (see, e.g., Figure~\ref{fig:proposedFramework}), hence the concatenation of 2D+3D features leads to a 1024-dimensional feature vector.  Figure~\ref{fig:embedding} illustrates a 2D embedding of the texture features, the depth features, and the combination of both. The 2D mapping is learned by optimizing the t-SNE \cite{maaten2008visualizing} over the train partition, then a random subset of test subjects are mapped for visualization. As we can see, 3D features favor the compactness and increase the distance between clusters associated to different subjects. 

To test the recognition performance, the images of the test subjects are partitioned into two sets: gallery and probe. For all the images in both sets, the 2D and 3D feature embedding is computed (using the pre-trained networks described before). Then, for each image in the probe set, the $n$ nearest neighbors in the gallery set are selected. The distance between each sample (in the embedding space) is measured using the distance defined in Section~\ref{sec:ProposedApproach}, Equation~\eqref{eq:distance}. For each sample in the probe set, we consider the classification as accurate, if at least one of the $n$ nearest neighbors is a sample from the same subject. The Rank-n accuracy is the percentage of samples in the probe set accurately classified. 

\begin{figure}
    \centering\includegraphics[width=.6\columnwidth]{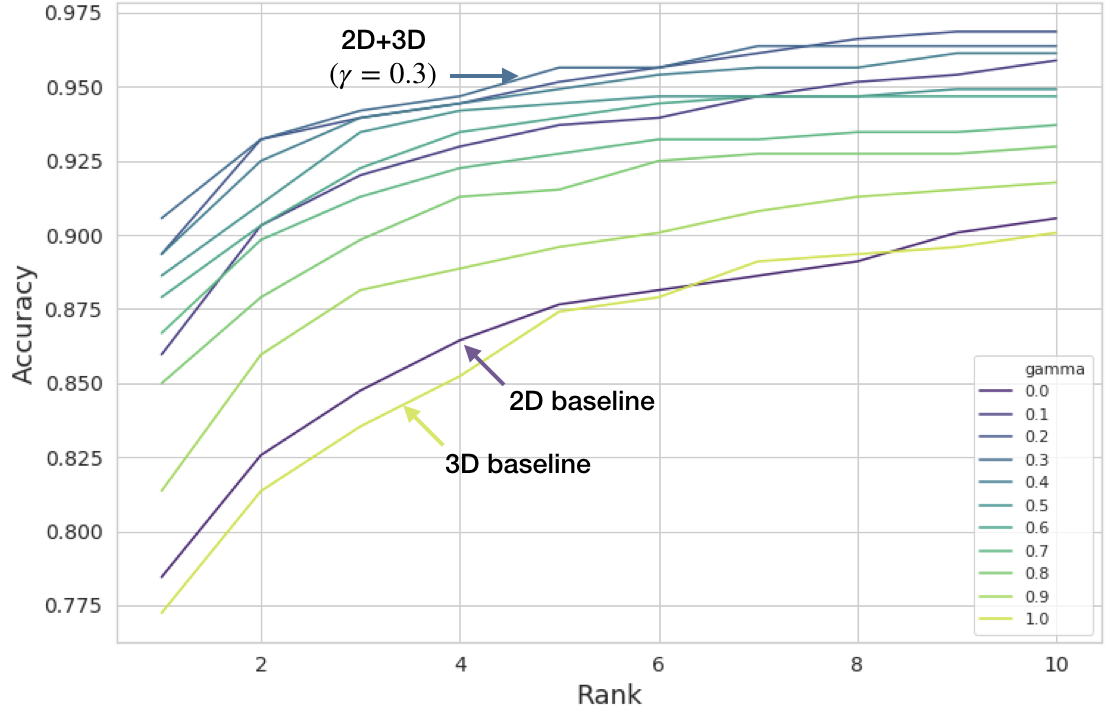}
    \caption{Rank-n accuracy for 2D, 3D, and 2D+3D face recognition. As discussed in Section~\ref{sec:ProposedApproach} the value of $\gamma$ can be set to weight texture and depth information in the classification decision. The extreme cases are $\gamma=0$ (only texture is considered) and $\gamma=1$ (only depth is considered). These extreme cases are illustrated in yellow and blue respectively, while intermediate solutions ($0<\gamma<1$) are presented in tones of green.} 
    \label{fig:gammarec}
\end{figure}
Figure~\ref{fig:gammarec} and Table~\ref{tab:Rank-n_rgb_vs_depth} show the Rank-n accuracy when: only 2D features ($\gamma=0$), only 3D features ($\gamma=1$), or a combination of both ($0<\gamma<1$) is considered. As explained in Section~\ref{sec:distanceDesign}, the value of $\gamma$ can be used to balance the weight of texture and depth features. As we can see, in all the cases a combination of texture and depth information outperforms each of them individually. This is an expected result as classification tends to improve when independent sources of information are combined \cite{Kuncheva2004}. $\gamma$ is an hyper-parameter that should be set depending on the conditions at deployment. In our particular experiments the best results are obtained for $\gamma=0.3$, which suggests that RGB features are slightly more reliable than depth features. This is an expected result as the module that extract RGB features is typically trained in a much larger datasets (2D facial images became ubiquitous). We believe this may change if, for example, testing is performed under low light conditions \cite{Lezama2017}. Testing this hypothesis is one of the potential path for future research. In the experiment discussed so far, we ignored the role of $\beta$ and $\alpha$ (i.e., we set $\beta=\infty$ and $\alpha=1$). As we will discuss in the following, these parameters become relevant to achieve jointly face recognition and spoofing prevention.
\setlength{\tabcolsep}{2pt}
\begin{table}
	\begin{center}
		\caption{Rank-n accuracy for 2D, 3D, and 2D+3D face recognition. As discussed in Section~\ref{sec:ProposedApproach} the value of $\gamma$ can be set to weight the impact of texture and depth information. The extreme cases are $\gamma=0$ (only texture is considered) and $\gamma=1$ (only depth is considered)}
		\label{tab:Rank-n_rgb_vs_depth}
		\begin{tabular}{L{3.2cm} C{1cm} C{1cm} C{1cm} C{1cm}}
			Rank-n Accuracy & 1 & 2 & 5 & 10 \\ 
			\hline\hline\noalign{\smallskip}
			RGB baseline ($\gamma=0$)       & $78.5$ & $82.6$ & $87.7$ & $90.6$ \\
			Depth baseline ($\gamma=1$)     & $77.2$ & $81.4$ & $87.4$ & $90.1$ \\
			(our) $\gamma=0.3$   & $\bf{90.6}$ & $\bf{93.2}$ & $\bf{95.6}$ & $\bf{96.4}$ \\
			(our) $\gamma=0.5$              & $88.6$ & $91.0$ & $94.4$ & $94.9$ \\
			(our) $\gamma=0.8$              & $85.0$ & $87.9$ & $91.5$ & $93.0$ \\
			\hline
		\end{tabular}
	\end{center}
\end{table}

\paragraph{Robustness to spoofing attacks.}
Spoofing attack are simulated to test face recognition models, in particular, how robust these frameworks are under (unseen) spoofing attacks. As in the present work we focus on the combination of texture and depth based features, the simulation of spoofing attacks must account for realistic texture and depth models. The models for the synthesis of spoofing attacks are described in detail in the supplementary material Section~\ref{app:implementation_spoofing}. 

\begin{figure}[htp]
    \centering
    \includegraphics[width=\columnwidth]{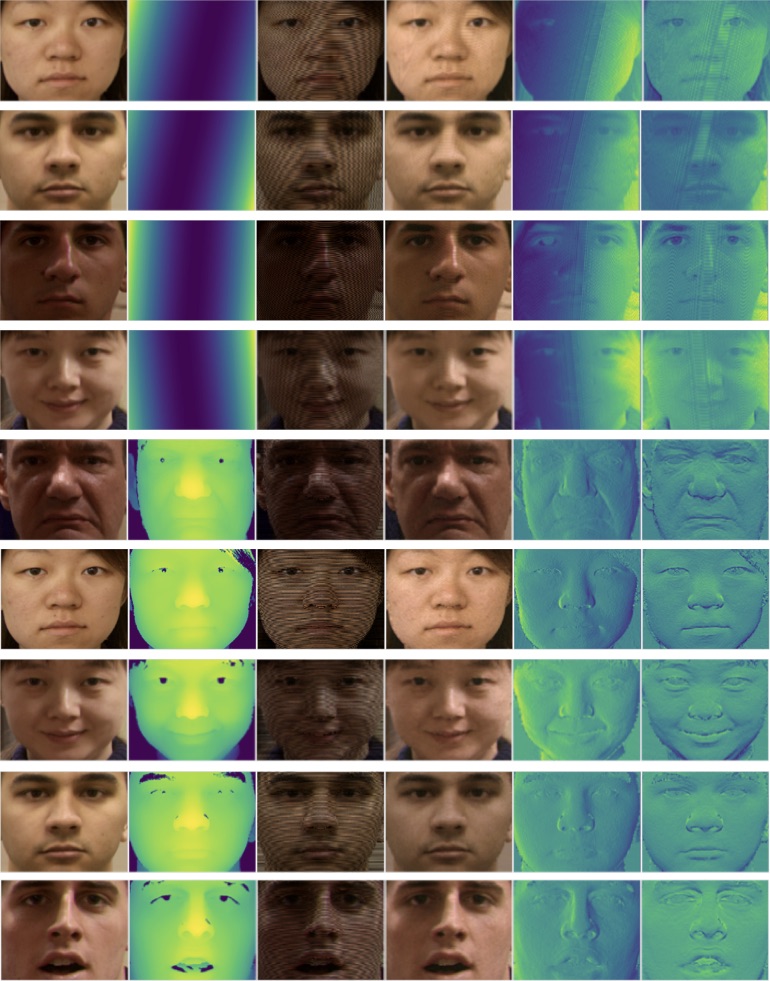}
    \caption{Examples of samples from live subjects and spoofing attacks. From left to right: (1) the ground truth texture, (2) the ground truth depth, (3) the input to our system (image with the projected pattern), (4) the recovered texture component (one of the outputs of the decomposition network), (5)/(6) recovered $x$/$y$ depth partial derivative. The first four rows correspond to spoofing samples (as explained in Section~\ref{app:implementation_spoofing}), and the bottom five rows to genuine samples from live subjects.}
    \label{fig:more_spoofing_examples}
\end{figure}
Figure~\ref{fig:more_spoofing_examples} illustrates spoofing samples (first four rows) and genuine samples (bottom five rows). The first two columns correspond to the ground truth texture and depth information, the third column illustrates the input to our system, and the last three columns correspond to the outputs of the decomposition network. These three last images are fed into the feature extraction modules for the extraction of texture and depth based features respectively, as illustrated in Figure~\ref{fig:proposedFramework}. It is extremely important to highlight, that spoofing samples are included exclusively at testing time. In other worlds, during all the training process the entire framework is agnostic to the existence of spoofing examples. If the proposed framework is capable of extracting real 3D facial features, it should be inherently robust to most common types of spoofing attacks. 

\begin{figure}[htp]
    \centering
    \includegraphics[width=.9\columnwidth]{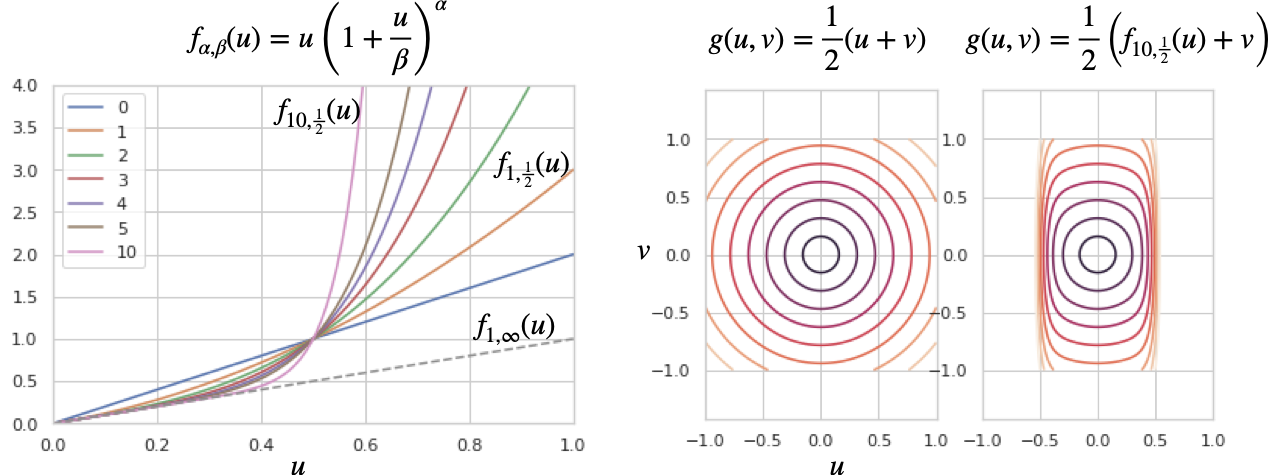}
    \caption{Illustration of the properties of the distance function defined in \eqref{eq:distance}. On the left side we illustrate the role of the parameter $\alpha$, and on the right, we compare the proposed distance and the standard euclidean distance. As can be observed, both measures are numerically equivalent in the region $[-\beta/2,\beta/2]\times[-\beta/2,\beta/2]$, but the proposed measure gives a higher penalty to vectors whose $u$ coordinate exceeds the value $\beta$.}\label{fig:distance_illustration}
\end{figure}

\begin{figure}[htp]
    \centering
    \includegraphics[width=.6\columnwidth]{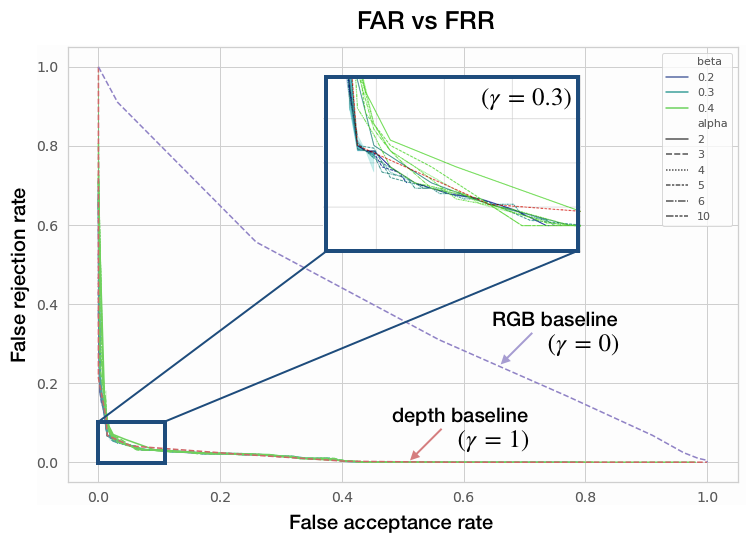}
    \caption{False acceptance rate and false rejection rate under the presence of spoofing attacks. On color blue we illustrate the RGB baseline ($\gamma=0$), on the other extreme, the red curve illustrates the performance when only depth features are considered. The combination of RGB and depth features is illustrated in tones of green for different values of $\alpha$ and $\beta$ (in this experiment we set $\gamma=0.3$).}
    \label{fig:spoofing_roc}
\end{figure}

As discussed before, the combination of texture and depth based features improves recognition accuracy. On the other hand, when spoofing attacks are included, we observe that texture based features are more vulnerable to spoofing attacks (see for example figure~\ref{fig:more_spoofing_examples} and \ref{fig:spoofing_roc}). To simultaneously exploit the best of each feature component, we design a non-linear distance as described in Equation~\eqref{eq:distance}. Figure~\ref{fig:distance_illustration} illustrates the properties of the defined distance for different values of $\alpha$ and $\beta$. As it can be observed, for those genuine samples (relative distances lower than $\beta$) the non linear component can be ignored and the distance behave as the euclidean distance with a relative modulation set by $\gamma$. On the other hand, if the distance between the depth components is above the threshold $\beta$, it will dominate the overall distance achieving a more robust response to spoofing attacks.  

To quantitatively evaluate the robustness against spoofing attacks, spoofing samples are generated for all the subjects in the test set. As before, the test set is separated into a gallery and a probe set and the generated spoofing samples are aggregated into the probe set. For each image in the probe set, the distance to a sample of the same subject in the gallery set is evaluated. If this distance is below a certain threshold $\lambda$, the image is labeled as genuine, otherwise, the image is labeled as spoofing. Comparing the classification label with the ground truth label we obtain the number of true positive (genuine classified as genuine), false positive (spoofing classified as genuine), true negative (spoofing classified as spoofing), and false negative (genuine classified as spoofing). Changing the value of the threshold $\lambda$ we can control the number of false positive versus the number of false negatives as illustrated in Figure~\ref{fig:spoofing_roc}.

Figure~\ref{fig:spoofing_roc} shows the ratio of false positive and false negative for $\lambda\in[0,2]$. As before the distance between the samples is computed using the definition provided in \eqref{eq:distance}, in blue/red the RGB/depth baseline is illustrated, the other set of curves (displayed in green tones) correspond to a combination of texture and depth features with $\gamma=0.3$ and different values of $\alpha$ and $\beta$. 
In Table~\ref{tab:spoofing_tpr} the ratio of true positive is reported for a fixed ratio of false positives. The ACER measure (last column) corresponds to the average between the ratio of spoofing and genuine samples misclassified.  
\setlength{\tabcolsep}{2pt}
\begin{table}
	\begin{center}
		\caption{Spoofing detection results. The ratio of true positive for a fixed ratio of false positive and the ACER measure are reported. Texture and depth facial features are combined using the distance defined in \eqref{eq:distance}. As we can see, the parameters $\gamma,\ \alpha$, and $\beta$ can be set to obtain better facial recognition performance and robustness against spoofing detection.}
		\label{tab:spoofing_tpr}
		\begin{tabular}{L{4.95cm}C{1.8cm} C{1.8cm} C{1.5cm}}
			  & TPR\% @FPR=$10^{-3}$ & TPR\% @FPR=$10^{-2}$ & ACER \% \\
			\hline\hline\noalign{\smallskip}
			RGB baseline ($\gamma=0$)                    & $21.8$      & $24.0$      & $38.9$ \\
			Depth baseline ($\gamma=1$)                  & $\bf{88.4}$ & $\bf{97.1}$ & $4.0$ \\
			(our) $\gamma=0.3,\ \beta=0.35\ \alpha=2$    & $85.5$      & $96.9$      & $4.5$ \\
			(our) $\gamma=0.3,\ \beta=0.35\ \alpha=5$    & $83.8$      & $\bf{97.1}$ & $4.0$ \\
			(our) $\gamma=0.3,\ \beta=0.35\ \alpha=10$   & $85.0$      & $95.6$      & $\bf{3.9}$ \\
			(our) $\gamma=0.3,\ \beta=0.4\ \alpha=2$     & $82.6$      & $96.9$      & $4.7$ \\
			(our) $\gamma=0.3,\ \beta=0.4\ \alpha=5$     & $86.4$      & $\bf{97.1}$ & $4.4$ \\
			(our) $\gamma=0.3,\ \beta=0.4\ \alpha=10$    & $81.8$      & $\bf{97.1}$ & $4.1$ \\
			(our) $\gamma=0.3,\ \beta=0.5\ \alpha=2$     & $86.4$      & $96.4$      & $5.3$ \\
			(our) $\gamma=0.3,\ \beta=0.5\ \alpha=5$     & $82.8$      & $95.6$      & $5.7$ \\
			(our) $\gamma=0.3,\ \beta=0.5\ \alpha=10$    & $85.0$      & $94.4$      & $5.9$ \\
			\hline
		\end{tabular}
	\end{center}
\end{table}

\paragraph{Testing variations on the ambient illumination.}
To test the impact of variations on lighting conditions we simulated test samples under different ambient illumination, implementation details are described in the supplementary material Section~\ref{app:simulating_lighting_conditions}. Table~\ref{tab:diff_light_acc} compares the rank-5 accuracy of 2D features and 2D+3D features as the power of the ambient illumination increases. As described in the supplementary material, the ambient illumination is modeled with random orientation, and therefore, the more powerful the illumination is the more diversity between the test and the gallery samples is introduced. 
\setlength{\tabcolsep}{2pt}
\begin{table}
	\begin{center}
		\caption{Recognition accuracy under different ambient illumination conditions. The power of the additional ambient light is provided relative to the power of the projected light, i.e., power=200\% means that the added ambient illumination is twice as bright as the projected pattern.}
		\label{tab:diff_light_acc}
		\begin{tabular}{L{4.95cm}C{2.0cm} C{2.0cm} C{2.0cm}}
			Rank-5 Accuracy  & power=100\% & power=150\% & power=200\% \\
			\hline\hline\noalign{\smallskip}
			RGB baseline ($\gamma=0$)          & $89.2$      & $81.2$      & $53.9$ \\
			(our) $\gamma=0.5$                 & $93.6$      & $90.7$      & $80.7$ \\
			\hline
		\end{tabular}
	\end{center}
\end{table}

In the present experiments, we assumed that both the projected pattern and the ambient illumination have similar spectral content. In practice, one can project the pattern, e.g., on the infrared band. This would make the system invisible to the user, and reduce the sensitivity of 3D features to variations on the ambient illuminations. We provide a hardware implementation feasibility study and illustrate how the proposed ideas can be deployed in practice in the supplementary material Section~\ref{app:hardware implementation}. 

\paragraph{Improving state of the art 2D face recognition.}
To test how the proposed ideas can impact the performance of state-of-the-art 2D face recognition systems, we evaluated our features in combination with texture based features obtained with ArcFace \cite{arcface}. ArcFace is a powerful method pre-trained on very large datasets, on ND-2006 examples it achieves perfect recognition accuracy (100\% rank-1 accuracy). When ArcFace is combined with the proposed 3D features, the accuracy remains excellent (100\% rank-1 accuracy), i.e., adding the proposed 3D features does not negatively affects robust 2D solutions. On the other hand, 3D features improve ArcFace on challenging conditions as we discuss in the following. Interesting results are observed when ArcFace is tested under spoofing attacks, as we show in Table~\ref{tab:spoofing_tpr_arcface}, ArcFace fails to detect spoofing attacks. ArcFace becomes more robust when it is combined with 3D features, improving from nearly $0$ TPR@FPR($10^{-3}$) to $84\%$. In summary, as 2D methods improve and become more accurate, our 3D features do not affect them negatively when they work well, while improve their robustness in challenging situations.
\setlength{\tabcolsep}{2pt}
\begin{table}
	\begin{center}
		\caption{Spoofing detection results for ArcFace and ArcFace enhanced with 3D features. Like in Table \ref{tab:spoofing_tpr}, the ratio of true positive for a fixed ratio of false positive and the ACER measure are reported.}
		\label{tab:spoofing_tpr_arcface}
		\begin{tabular}{L{4.95cm}C{2.0cm} C{2.0cm} C{1.5cm}}
			  & TPR\% @FPR=$10^{-3}$ & TPR\% @FPR=$10^{-2}$ & ACER \% \\
			\hline\hline\noalign{\smallskip}
			ArcFace ($\gamma=0$)          & $0$      & $0$      & $46.2$ \\
			ArcFace + 3D $(\gamma=0.5)$                           & $84.7$   & $94.7$   & $7.9$ \\
			\hline
		\end{tabular}
	\end{center}
\end{table}

\section{Conclusions}\label{sec:Conclusions}
We proposed an effective and modular alternative to enhance 2D face recognition methods with actual 3D information. 
A high frequency pattern is designed to exploit the high resolution cameras ubiquitous in modern smartphones and personal devices.
Depth gradient information is coded in the high frequency spectrum of the captured image while a standard texture facial image can be recovered to exploit state-of-the-art 2D face recognition methods. 
We show that the proposed method can be used to simultaneously leverage 3D information and texture information. 
This allows us to enhance state-of-the-art 2D methods improving their accuracy and making them robust, e.g., to spoofing attack. 


\section*{Acknowledgments}
Work partially supported by ARO, ONR, NSF, and NGA.

\bibliographystyle{unsrt}  
\bibliography{bibfile.bib}

\clearpage
\iftrue  
\input{supplementary.tex}
\fi  
\end{document}

%% file: supplementary.tex
\newpage{}
\begin{center}
  \textbf{\Large{Supplementary Material}}  
\end{center}

\appendix

\section{Limitations of 3D hallucination in the context of face recognition.}
As discussed in Section~\ref{sec:relatedWork} 3D hallucination methods have intrinsic limitations in the context of face recognition. To complement the example illustrate in Figure~\ref{fig:3dillustration}, here we show 5 3D facial models obtained by hallucinating 3D from a single RGB input image. To that end, we apply the 3D morphable model (extremely popular in facial applications). As we can see, even in the case of a planar spoofing attack, a face-like 3D shape is retrieved despite that this is far form the actual 3D shape of the actual scene. This is an expected results, as we discuss in Section~\ref{sec:relatedWork} the problem of 3D hallucination is ill-posed, and therefore priors need to be enforced in order to obtain feasible implementations.
\begin{figure}[htp]
    \centering
    \includegraphics[width=.8\columnwidth]{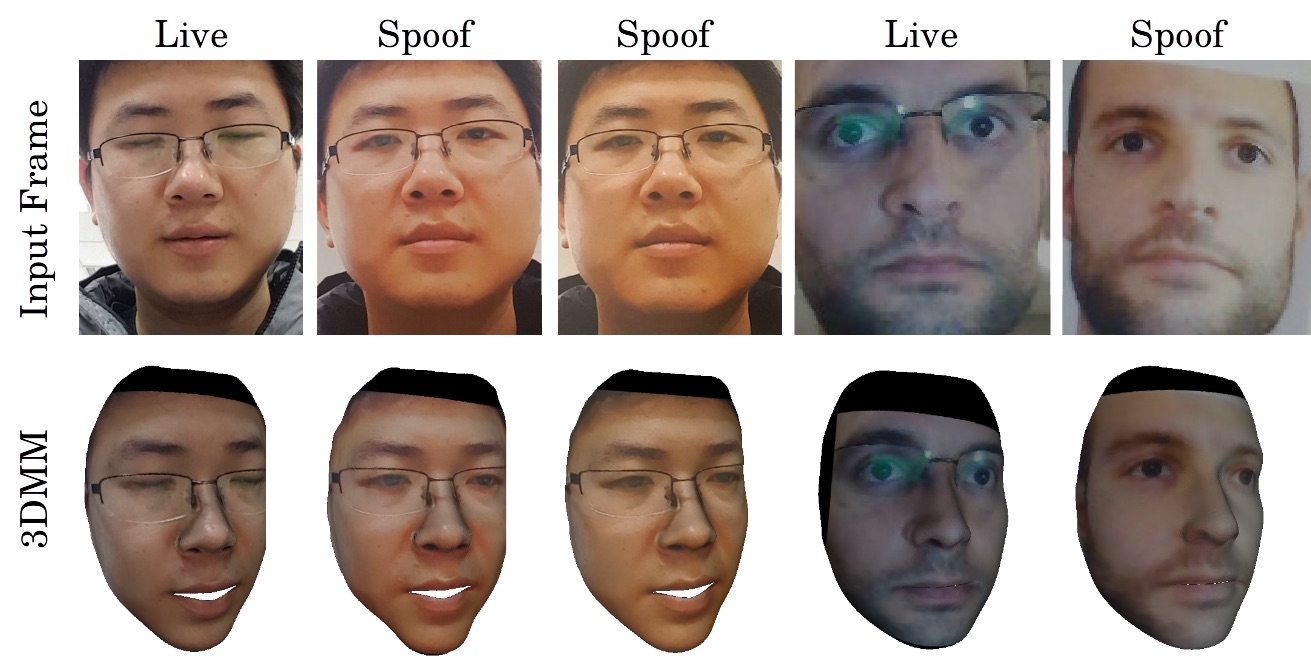}
    \caption{Example of 3D hallucination \cite{blanz2003face} from photographs of live faces versus portraits. Second row illustrates the 3D morphable model (second row) of images of live subjects (first and fourth columns) and for photographs of portraits of the same subjects (second, third and fifth columns). Results obtained using the code from \cite{Huber2015}}
    \label{fig:3dmm}
\end{figure}

\section{Review of Active Stereo Geometry}\label{sec:reviewActiveStereo}
\begin{figure}
\centering\includegraphics[width = .5\columnwidth]{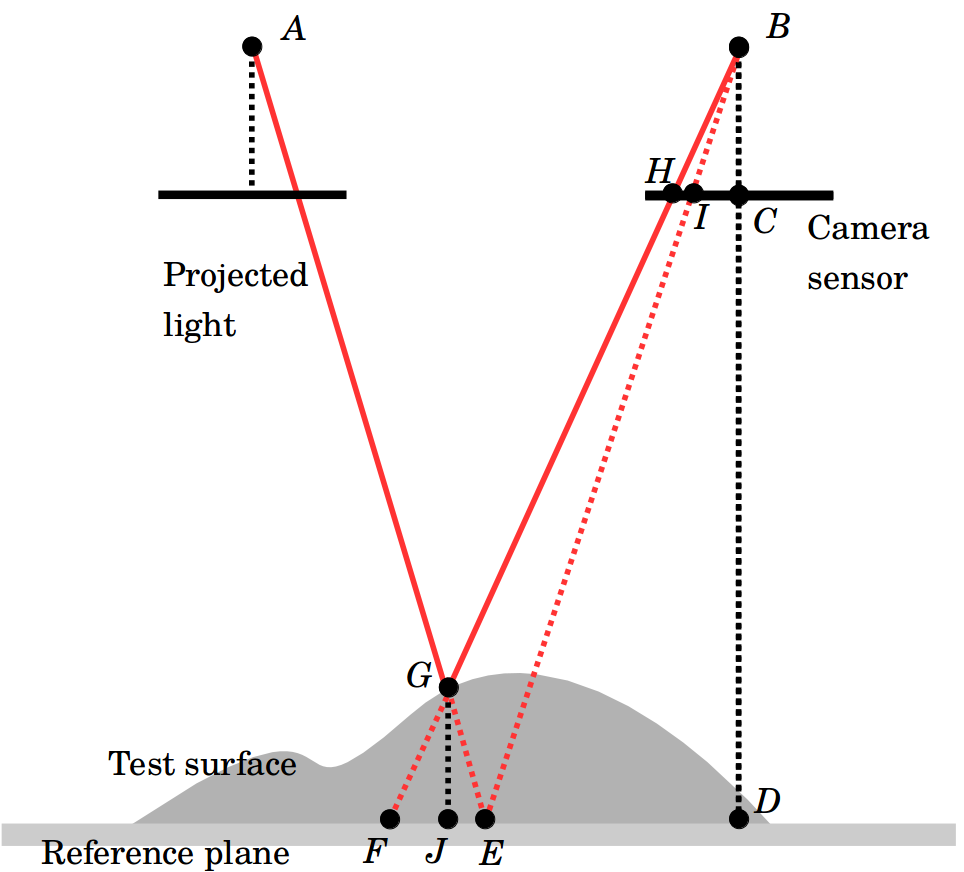}
\caption{Active stereo geometry. Left sensor illustrates the optical image plane of the device projecting light (with optical center at point $A$), and the right sensor illustrates the camera sensor (with optical center at $B$). A ray of light projected over a reference plane (at point $E$) is perceived by the camera at the pixel position $I$. When the same ray of light is projected over an arbitrary surface (at point $G$), it is perceived by the camera at a shifted location $H$. The disparity $\overline{IH}$ is proportional to $\overline{JG}$ (height of the surface).}\label{fig:active_stereo}
\end{figure}

When a structured pattern of light is projected over a surface, it is perceived with a certain deformation depending on the shape of the surface.
For example, the top left image in Figure~\ref{fig:fringesDescomposition} is acquired while horizontal stripes are projected over the subject face. 
As we can see, fringes are no longer perceived as horizontal and parallel to each other, this deformation codes rich information about the geometry of the scene (which will be exploited to enhance 2D face recognition methods).
Figure~\ref{fig:active_stereo} sketches the geometry of this situation as if we were looking at the face from the side (vertical section). 
On the left, a light source project a ray of light trough the points $AG$ (red line), when this ray is reflected by a reference plane at the point $E$, it is \emph{viewed} by the camera at the pixel location represented by $I$. 
If instead, light is projected over an arbitrary (non-planar) surface, the reflection is produced at point $G$ and viewed by the camera at the shifted position $H$. From now on, we denote the length of the shift $\overline{IH}$ as the disparity $d$. 

Similarity between triangles $\widehat{FBE}$-$\widehat{HBI}$ and $\widehat{FGE}$-$\widehat{AGB}$ leads to $\overline{EF}/\overline{DB} = \overline{IH}/\overline{CB}$ and $\overline{FE}/\overline{JG} = \overline{AB}/(\overline{DB} - \overline{BC})$. Defining $\overline{AB}=b$ (baseline between the lighting source and the camera sensor), $\overline{BC}=f$ (camera focal length), $\overline{JG}=z$ (surface local height), $\overline{BD} = q$ (distance of the subject to the camera), and assuming that $q \gg f$ we obtain,
\begin{equation}\label{eq:distarityvsdepth}
    \phi = z \frac{bf}{q^2} \Rightarrow \nabla \phi \propto \nabla z.
\end{equation}

Equation~\eqref{eq:distarityvsdepth} quantitatively relates the perceived shift (disparity) $d$ of the projected pattern and the surface 3D shape $z$. As the goal of the present work is to provide local 3D features instead of performing an actual 3D reconstruction, the particular value of $b$, $f$ and $h$ are irrelevant as we shall see. Moreover, we exploit the fact that the gradient of the local disparity codes information of the depth gradient. This allows us to extract local geometrical features bypassing the more challenging steps of 3D reconstruction: global matching and gradient field integration \cite{zhang2013handbook, di2018one, ghiglia1998unwrapping, hartley2003multiple}. 

\section{Gradient Information is Easy to Compute, Absolute Depth is Hard.}
One of the key ideas of the presented approach is to estimate local depth descriptors instead of absolute depth information. While the former is an easier task and absolute information is irrelevant for the sake of feature extraction. After all, a robust feature representation should be scale and translation invariant.  

More precisely, we can define the problem of integrating the absolute depth $z$ from an empirical estimation of its gradient map $(\tilde{z_x},\tilde{z_y})$ in a 2D domain $\Omega\in\mathbb{R}^2$ as the optimization problem:
\begin{equation}\label{eq:integration}
z^* = \text{argmin}_{z\in\mathcal{S}}\, \iint_\Omega \left\|\nabla z - (\tilde{z_x},\tilde{z_y})^T \right\| \, \partial\Omega.
\end{equation}

The solution of Equation~\eqref{eq:integration} is non trivial. Noise, shadows, and facial discontinuities produce empirical gradient estimations $(\tilde{z_x},\tilde{z_y})$ with irotational components, i.e., for some pixels $(x_i,y_i)$ the rotor of the field is different from zero $\nabla\times(\tilde{z_x}(x_i,y_i),\tilde{z_y}(x_i,y_i))^T \neq 0$. Therefore, the space of target functions $\mathcal{S}$ and the minimization norm $\| \cdot \|$ must be carefully set in order to achieve a meaningful solution to Equation~\eqref{eq:integration}. This is a complex mathematical problem and has been extensively studied in the literature \cite{agrawal2005algebraic, agrawal2006range, du2007robust, reddy2009enforcing, tumblin2005want}.

\section{Implementation details}\label{app:implementation_details}

\begin{figure}
    \centering
    \includegraphics[width=.6\columnwidth]{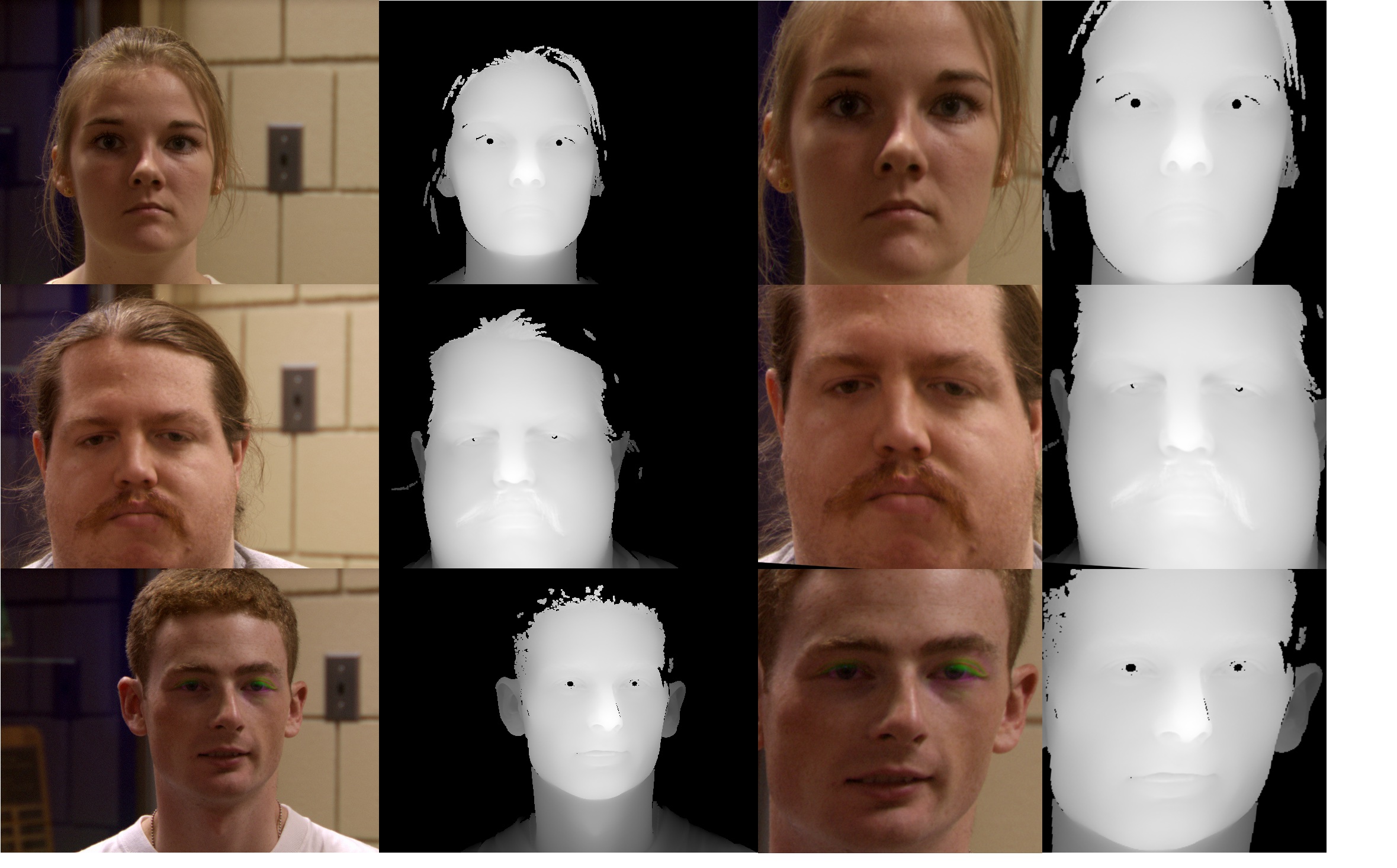}
    \caption{Samples from the nd2006 database. On the left column, the RGB images can be seen, while on the right their corresponding depth images are shown.} 
    \label{fig:nd2006samples}
\end{figure}

\subsection{Light projection}\label{app:sec:sim_proj_pat}
\begin{figure}[htp]
    \centering\includegraphics[width=.5\textwidth]{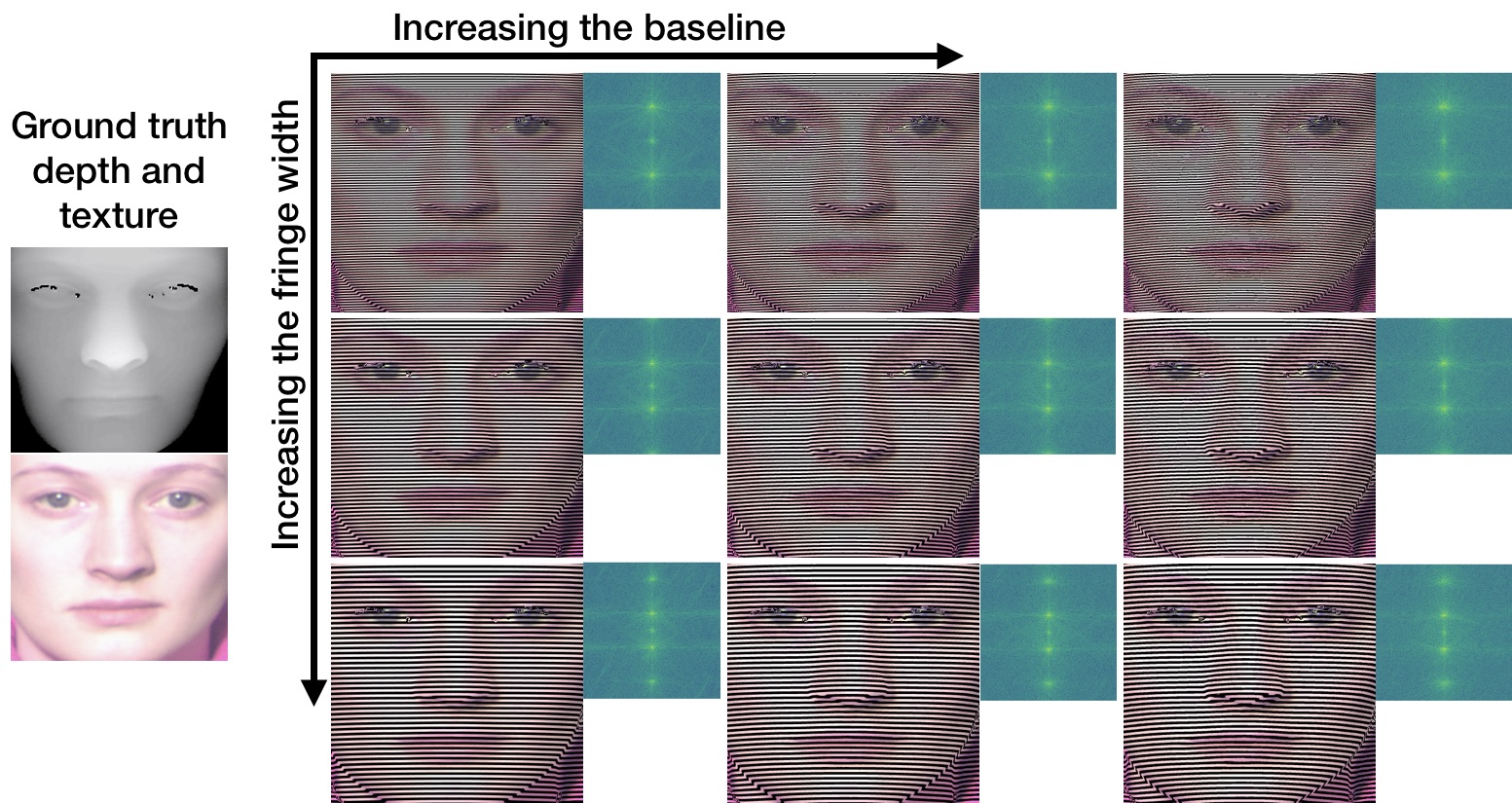}
    \caption{Texture and shape information on a single RGB image. On the left side we show the ground truth depth (top) and texture (bottom) facial information for a samples of ND-2006 dataset. With the geometric model described in Section~\ref{sec:reviewActiveStereo}, we simulated images acquired under the projection of a periodic fringe pattern. The absolute value of the Fourier transform (in logarithmic scale) is illustrated at the right side of each example. From left to right, we show who the baseline (distance between the light source and the camera) impact the simulation. Recall the role of the baseline, defined as $b$ in Equation~\ref{eq:distarityvsdepth}. From the top to the bottom, we show the effect of the width of the fringes, which define the fundamental frequency of the pattern $f_0=1/T$, see Section~\ref{sec:pattern design}. Note that as we are displaying high frequency patterns in the Figure, the reader may be observing aliasing artifacts due to a poor pdf resolution (zooming into the image is recommended).}
    \label{fig:generation_of_d3d}
\end{figure}
From ground truth depth and texture facial information, images under the projection different high frequency patterns can be simulated as illustrated in Figure~\ref{fig:generation_of_d3d}. Model the physical configuration of the system as described in Section~\ref{sec:reviewActiveStereo}, and different parameters for the baseline and the fringe width were tested. We assumed in all our experiments a fixed focal length. The pseudo-code for the generation of samples under active illumination is summarized in Algorithm~\ref{alg:generation_active_proj}. It is important to highlight that thought the problem of simulating the pattern deformation from the depth is easy, the opposite is a very hard problem. This is why we design a DNN-based approach that estimates from the deformation of the pattern the gradient of the depth rather than the depth itself (see Section~\ref{sec:ProposedApproach}). 
\begin{algorithm}
\begin{algorithmic}[1]
\Procedure{LightProjection}{$D(x,y), T(x,y)$}
\Statex Read the light pattern to be projected (pre-designed).
\State $p(x,y) = \mbox{loadLightProfile}()$
\Statex Simulate the local disparity (see~\eqref{eq:distarityvsdepth}).
\State $disparity(x,y) = \frac{bf}{q^2} D(x,y)$ 
\Statex Compute local pattern deformation.
\State $\tilde{p}(x,y) = \mbox{bilinearInterpolation}(p, (x+d(x,y),y))$
\Statex Account for the texture.
\State $O(x,y) = \tilde{p}(x,y)\, T(x,y)$
\State \textbf{return} $O(x,y)$ \Comment{Image with active illumination.}
\EndProcedure
\end{algorithmic}
\caption{Active light projection. Model the resulting RGB image when a pattern of structured light $p(x,y)$ is projected over a surface with depth profile $D(x,y)$ and texture $T(x,y)$.}\label{alg:generation_active_proj}
\end{algorithm}

\subsection{Networks architecture} \label{sec:decompositionNet} \label{sec:embeddingNets}
Table~\ref{tab:DescompositionNetwork} illustrates the architecture of the network that performs texture and depth information decomposition (as described in Section~\ref{sec:ProposedApproach}). Table~\ref{tab:FeatureEmbeddingNetwork} illustrates the architecture of the layers trained for facial feature extraction (illustrated as yellow/green block in Figure~\ref{fig:proposedFramework}). Each module of the proposed framework is implemented using standard tensorflow layers (version 1.13). 
\begin{table}[ht!]
	\begin{center}
		\caption{Decomposition network (illustrated in blue in Figure~\ref{fig:proposedFramework}). Conv denotes convolution layer, and BN batch normalization. Standard tensorflow (v1.13) layers are used.} \label{tab:DescompositionNetwork} 
		\begin{tabular}[t]{ll}
			\hline\noalign{\smallskip}
			\textbf{Input:} & $480{\times}480 {\times} 3$ image with proj. fringes. \\ \arrayrulecolor{gray}\hline
			(A) RGB - branch \\ \hline 
			Layer A.1: & Conv - $64$ kernels $4{\times}4$ , BN, LeakyRelu. \\
			Layer A.2: & Conv - $4$ kernels $3{\times}3$ , BN, LeakyRelu. \\
			Layer A.3: & Conv - $4$ kernels $3{\times}3$ , BN, LeakyRelu. \\
			Layer A.4: & Conv - $2$ kernels $1{\times}1$ , BN, LeakyRelu. \\
			Layer A.5: & Resize($112{\times}112 {\times} 3$ \\
			\textbf{Output A:} & $112{\times}112 {\times} 3$ recovered texture. \\ 
            \arrayrulecolor{gray} \hline 
			(B) Depth - branch \\ \hline 
			Layer B.1: & Average(axis=3) (convert to gray). \\
			Layer B.2: & Conv - $64$ kernels $4{\times}4$ , BN, LeakyRelu. \\
			Layer B.3: & Conv - $4$ kernels $3{\times}3$ , BN, LeakyRelu. \\
			Layer B.4: & Conv - $4$ kernels $3{\times}3$ , BN, LeakyRelu. \\
			Layer B.5: & Conv - $2$ kernels $1{\times}1$ , BN, LeakyRelu. \\
			Layer B.6: & Resize($112{\times}112 {\times} 2$ \\
			\textbf{Output B:} & $112{\times}112 {\times} 2$ recovered $\{z_x, z_y\}$. \\ \arrayrulecolor{black}\hline
		\end{tabular}
	\end{center}
\end{table}
\begin{table}[ht!]
	\begin{center}
		\caption{Feature embedding (illustrated in yellow/green in Figure~\ref{fig:proposedFramework}). Conv denotes convolution layer, SepConv separable convolution, BN batch normalization, IN instance normalization. Standard tensorflow (v1.13) layers are used.} \label{tab:FeatureEmbeddingNetwork} 
		\begin{tabular}[t]{ll}
			\hline\noalign{\smallskip}
			\textbf{Input:} & $112{\times}112 {\times} c$ ($c=3$ for texture, $c=2$ for depth image). \\ \arrayrulecolor{gray}\hline
			Layer 1.1: & Conv - $32$ kernels $3{\times}3$ stride $2$, IN, Relu. \\
			Layer 1.2: & Conv - $64$ kernels $3{\times}3$ stride $2$, IN, Relu. \\
			\arrayrulecolor{gray}\hline
			\textbf{Path A1} & \\
			Layer A1.1: & Conv - $128$ kernels $1{\times}1$ stride 2, IN. \\
			\arrayrulecolor{gray}\hline
			\textbf{Path B1} & \\
			Layer B1.1: & SepConv - $128$ kernels $3{\times}3$, IN, Relu. \\
		    Layer B1.2: & SepConv - $128$ kernels $3{\times}3$, IN, Relu. \\
			Layer B1.3: & MaxPooling $2{\times}2$. \\
			\arrayrulecolor{gray}\hline
			Layer 2: & Path A1 + Path B1 (output B1.3 + output A1.1) \\
			\arrayrulecolor{gray}\hline
			\textbf{Path A2} & \\
			Layer A2.1: & Conv - $384$ kernels $1{\times}1$ stride 2, IN. \\
			\arrayrulecolor{gray}\hline
			\textbf{Path B2} & \\
			Layer B2.1: & SepConv - $384$ kernels $3{\times}3$, IN, Relu. \\
		    Layer B2.2: & SepConv - $384$ kernels $3{\times}3$, IN, Relu. \\
			Layer B2.3: & MaxPooling $2{\times}2$. \\
			\arrayrulecolor{gray}\hline
			Layer 3: & x = Path A2 + Path B2 (output B2.3 + output A2.1) \\
			\arrayrulecolor{black}\hline 
			\textbf{Middle flow:} & x: $28\times28\times 384$ (Repeat 2 times) \\
			Layer 4.1: & SepConv - $384$ kernels $3{\times}3$, IN, Relu. \\
		    Layer 4.2: & SepConv - $384$ kernels $3{\times}3$, IN, Relu. \\
		    Layer 4.3: & SepConv - $384$ kernels $3{\times}3$, IN, Relu. \\
            Layer 4.4: & Add(x, output Layer $4.3$) \\
			\arrayrulecolor{black}\hline
			\textbf{Exit flow:} & $28\times28\times 256$  \\
			\arrayrulecolor{gray}\hline
			\textbf{Path A5} & \\
			Layer A5.1: & Conv - $512$ kernels $1{\times}1$ stride 2, IN. \\
			\arrayrulecolor{gray}\hline
			\textbf{Path B5} & \\
			Layer B5.1: & SepConv - $256$ kernels $3{\times}3$, IN, Relu. \\
		    Layer B5.2: & SepConv - $512$ kernels $3{\times}3$, IN, Relu. \\
			Layer B5.3: & MaxPooling $2{\times}2$. \\
			\arrayrulecolor{gray}\hline
			Layer 6: & Path A5 + Path B5 (output B5.3 + output A5.1) \\
			Layer 7: & SepConv - $512$ kernels $3{\times}3$, IN, Relu. \\
			Layer 8: & SepConv - $512$ kernels $3{\times}3$, IN, Relu. \\
			Layer 9: & Global Average Pooling 2D.  \\
			Output:  & $512 \times 1$ facial features. \\
			\arrayrulecolor{black}\hline
		\end{tabular}
	\end{center}
\end{table}

\paragraph{Training protocol.} 
The training procedure consists of three phases, first we perform 10 epochs using stochastic gradient descend (SGD), then, we iterate 20 additional epochs using adam optimizer, and finally, we perform 10 epoch using SGD. During these phases the learning rate is set to $10^{-3}$. These three steps are commonly refer in the literature as ``network warmup'', ``training'', and ``fine-tunning'', we observed that training each framework module following this protocol leads to stable and satisfactory results (as reported in Section~\ref{sec:Experiments}). However, we did not focus in the present work on the optimization of the networks architecture, nor the training protocols.

\subsection{Simulation of spoofing attacks} \label{app:implementation_spoofing}
Spoofing attack are simulated to test face recognition models, in particular, how robust these frameworks are under (unseen) spoofing attacks. As in the present work we focus on the combination of texture and depth based features, the simulation of spoofing attacks must account for realistic texture and depth models. Algorithm~\ref{alg:synthesis_of_spoofing} summarizes the main steps for the simulation of spoofing attacks (which are detailed next).
%
\begin{algorithm}
\begin{algorithmic}[1]
\Procedure{TransformText}{$T_0$} \Comment{Sim. spoofing texture}
\Statex Init. optimal filter (this is done only once and off-line).
\State $k(x,y) = $ InitKernel(RealEx., SpoofEx.) \Comment{\eqref{eq:definition_spoofing_kernel}}
\Statex Filter the genuine sample.
\State $T(x,y) = $ filter$(T_0, k(x,y))$
\State \textbf{return} $T(x,y)$ \Comment{Simulated spoofing texture}
\EndProcedure
\Procedure{TransformDepth}{} \Comment{Sim. spoofing depth}
\Statex Compute spoofing depth
\State $z(x,y)$ = SpoofDepthModel() \Comment{E.g., \eqref{eq:depth_profile},\eqref{eq:depth_profile2}}. 
\State \textbf{return} $z(x,y)$ \Comment{Simulated spoofing depth}
\EndProcedure
\Procedure{SpoofingSample}{$\{T_0, z_0\}$}
\Statex Simulate sample texture.
\State $T(x,y) = $ TransformText($T_0$)
\Statex Simulate sample depth.
\State $z(x,y) = $ TrasformDepth()
\State \textbf{return} $\{T, z\}$ 
\EndProcedure
\end{algorithmic}
\caption{Synthesis of the texture and depth of spoofing attacks.}\label{alg:synthesis_of_spoofing}
\end{algorithm}

\begin{figure}[htp]
    \centering\includegraphics[width = .65\textwidth]{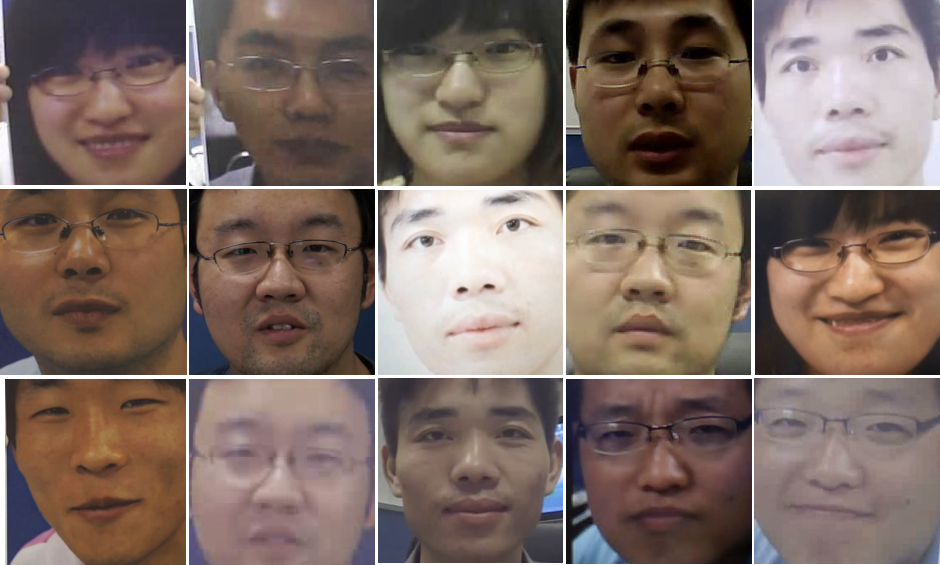}
    \caption{Spoofing and genuine examples from CASIA \cite{zhang2012face} dataset.}
    \label{fig:CASIA}
\end{figure}

To simulate realistic texture conditions, CASIA dataset is analyzed \cite{zhang2012face}. This dataset contains samples of videos collected under diverse spoofing attacks, e.g., video and photo attacks. 50 different subjects participated in the data collection, and 600 video clips were collected. We extracted a sub set of random frames from these videos (examples are illustrated in Figure~\ref{fig:CASIA}), and use them to identify certain texture properties that characterize spoofing attacks, as we explain next. 
 
Several works have been published in the resent years supporting that certain texture patterns are characteristics of spoofing photographs \cite{Atoum2017, Li2018, li2004live, Li2017spoofing, Liu2018spoofing, yeh2017face}. In particular, differences in the Fourier domain have been reported \cite{li2004live}, which provide cues for certain classes of spoofing attacks \cite{yeh2017face}. Following these ideas, we propose a simple model for the synthesis of spoofing attacks from genuine samples extracted from ND-2006 dataset. We assume a generic linear model $I_{spoof}(x,y) = I_{real}(x,y) \ast k(x,y)$, where $I_{spoof}$ represents the texture of the simulates spoofing attack for the genuine sample $I_{real}$ and $k$ an arbitrary kernel that will be set. 

Let us denote as $g_i(x,y)$ ($i=1 ,..., M$) a set facial images associated to spoofing attacks (here extracted from CASIA dataset), and $q_i(x,y)$ ($i=1,...,N$) a set of genuine facial images (for example, from CASIA or ND-2006 datasets). As in Section~\ref{sec:ProposedApproach} we denote $\tilde{q}(f_x,f_y)=\mathcal{F}\{q(x,y)\}$ the 2D discrete Fourier transform of $q(x,y)$. Given ground truth examples of genuine and spoofing facial images ($q_i$ and $g_i$), we define the kernel $k$ as 
\begin{equation}\label{eq:definition_spoofing_kernel}
    k(x,y) = \mathcal{F}^{-1}\left\{ \frac{Q(f_x,f_y)}{G(f_x,f_y) + \epsilon} \right\}, \ \mbox{s.t. } k(-x,-y) = k(x,y),
\end{equation}
where $G$ and $Q$ are defined as follow 
\begin{equation}\label{eq:define_G_and_Q}
    G = \frac{1}{M} \sum_{i=1}^{M} \|\tilde{g}_i(f_x, f_y)\|, \  Q = \frac{1}{N} \sum_{i=1}^{N} \|\tilde{q}_i(f_x, f_y)\|. 
\end{equation}
We set $\epsilon = 10^{-6}$ for numerical stability. Observe that $G$ and $Q$ account only for the absolute value of the Fourier transform of real and spoofing samples, while the phase information is discarded. In principle, an arbitrary phase factor can be included, we constrain the solution to be a symmetric kernels, which is equivalent to enforce a null phase. To perform the average defined in Equation~\eqref{eq:define_G_and_Q}, the 2D coordinates associated to the frequency domain must be refer to a common coordinate frame, this allows to aggregate the frequency information of images of heterogeneous resolution. 

\begin{figure}[htp]
    \centering
    \includegraphics[width = .65\textwidth]{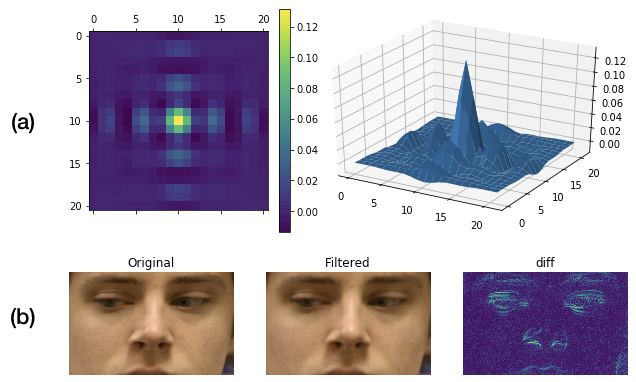}\hspace{2mm}
    \caption{Texture simulation of spoofing attacks. (a) Illustrates the linear kernel optimized such that the Fourier domain of real samples match that of spoofing samples. (b) Shows (left) an example of an image of a genuine subject (from ND-2006 set), the simulated texture of a spoofing attack (center), and finally (right) the different between the two.}
    \label{fig:simulation_of_spoofing_attacks_examples}
\end{figure}
Figure~\ref{fig:simulation_of_spoofing_attacks_examples}-(a) shows the kernel obtained using 600 spoofing samples from CASIA dataset and 600 genuine samples from ND-2006.
As we can see, the resulting kernel is composed of predominantly positive values, this suggest that the application of it will produce essentially a blurred version of the original image.
This empirical result is in accordance with the findings reported in \cite{li2004live, yeh2017face}. 
Figure~\ref{fig:simulation_of_spoofing_attacks_examples}-(b) shows an example of an image of a genuine face (left), the result of applying the estimated kernel $k$ (center) and the different between them (right). 

\begin{figure}[htp]
    \centering
    \includegraphics[width = .65\textwidth]{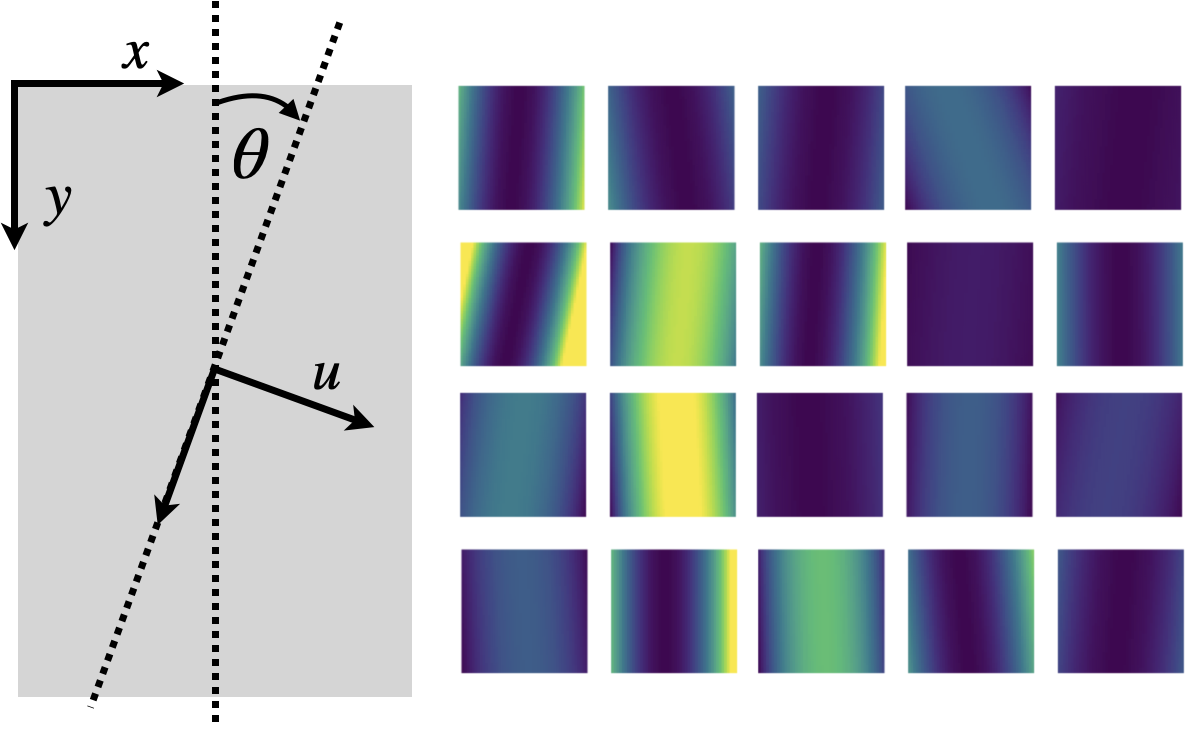}
    \caption{Depth simulation of the spoofing attacks. On the left we illustrate the geometry of the simulation process. The right side illustrates examples of the generated depth profiles associates to spoofing attacks.}
    \label{fig:depth_spoofing_simulation}
\end{figure}

Finally, the depth profile of different types of spoofing attacks is simulated. To this end, different kinds of polynomial forms were evaluated. We simulated planar attacks, which take place when the attack is deployed using a phone or a tablet, and non-planar attacks, common when the attacker uses a curved printed portrait. We modeled the latter as random parabolic 3D shapes. The principal axis of each surface is randomly oriented, presenting an angle $\theta\sim\mathcal{N}(0,(1/10\cdot\pi/2)^2)$ with respect to the vertical (as illustrates Figure~\ref{fig:depth_spoofing_simulation}). The depth profile z(x,y) is given by 
\begin{equation}\label{eq:depth_profile}
    z(x,y) = \frac{4a}{w^2}\left(u(x,y) - c \frac{w}{2}\right)^2
\end{equation}
where $u(x,y) = x\,\cos(\theta) + y\, \sin(\theta)$, $w$ is a constant representing the width of the spatial domain (here 480 pixels), $a$ sets the curvature of and is a random variable sampled from the normal distribution $\mathcal{N}(0, (2/10)^2)$, and $c$ (also a random variable) sets the offset of the principal axis and is sampled from the distribution $\mathcal{N}(0,(1/10)^2)$. We also tested arbitrary polynomial forms with random coefficients, e.g., 
\begin{equation}\label{eq:depth_profile2}
    z(x,y) = \sum_{\substack{i=0,..,3 \\ j=0,..,3}} w_{ij}\, a_{ij} {(x/w-b_{ij})}^i {(y/h-c_{ij})}^j
\end{equation}
where the coefficients are samples from an uniform distribution $a_{ij}, b_{ij}, c_{ij} \sim \mathcal{U}_{[0,1]}$, and $w_{ij} = 10^{\max{(0,i-1)}\max{(0,j-1)}}$ is a normalization factor.  
Figure~\ref{fig:depth_spoofing_simulation} illustrates (right side) some examples of spoofing depth profiles generated. 

\subsection{Facial appearance under different illumination} \label{app:simulating_lighting_conditions}
Using the original texture and depth facial information, the appearance of the face under novel illumination conditions can be simulated. One of the simplest and more accepted models consists of considering the facial surface as a lambertinan surface \cite{basri2003lambertian, shashua1997, Zhou2007}. Hence, the amount of light reflected can be estimated as proportional to the cosine of the angle between the surface local normal and the direction in which the light approaches the surface. 

More precisely, 
\begin{equation}\label{eq:reflectanceFace1}
I_{(x,y,z)} = a_{(x,y,z)}\, \max\left(\vec{n}_{(x,y,z)} \cdot p_0\vec{l},\ 0\right),
\end{equation}
where $I$ denotes the intensity of the light reflected, $p_0$ the intensity of the incident light, $a$ represents the surface albedo, $\vec{n}$ is a unit vector normal to the surface, and $\vec{l}$ a unit vector indicating the direction in which the light rays approach the surface at $(x,y,z(x,y))$.
If the depth profile $z(x,y)$ is know, the normal vector to the surface at $(x,y,z)$ can be computed as
\begin{equation}\label{eq:normal_vectors}
\vec{n}(x,y) = \frac 1 {\sqrt{\left(\dpar{z}{x}\right)^2 + \left(\dpar{z}{y}\right)^2 + 
1}} \left(-\dpar{z}{x}, -\dpar{z}{y}, 1\right)^t.
\end{equation}

\begin{figure*}[htp]
    \centering
    \includegraphics[width=.95\textwidth]{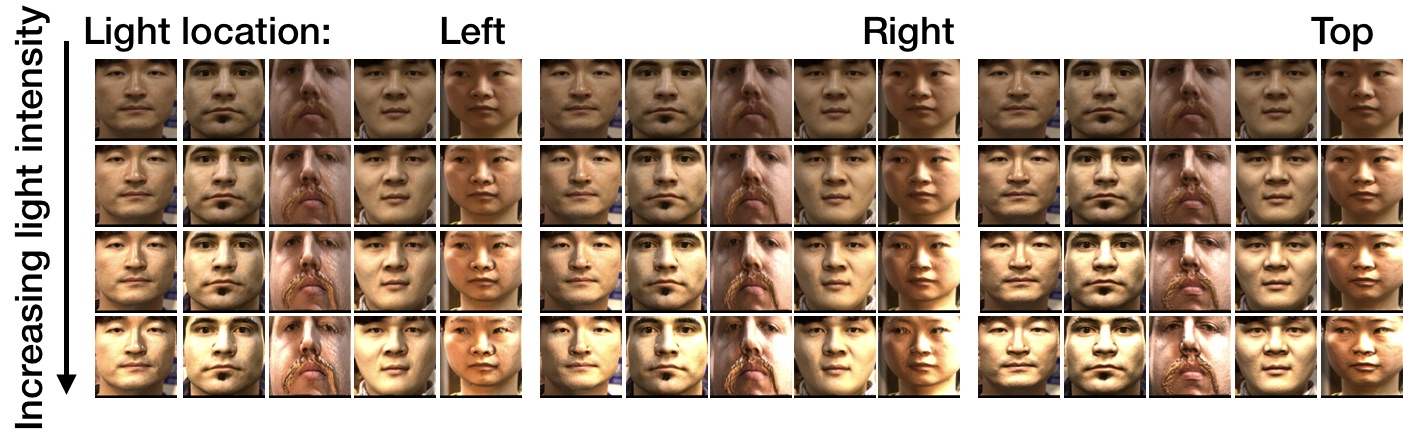}
    \caption{Examples of new samples under different lighting conditions. The groups on the left, middle and right, correspond to a light source located at the left, right, and top of the scene respectively. The images on the bottom are created assuming a brighter light source (higher value of $p_0$, see Algorithm~\ref{alg:diff_illumination}).}
    \label{fig:LightSimulation}
\end{figure*}

Algorithm~\ref{alg:diff_illumination} summarizes the main steps for simulating new samples under novel illumination conditions. Figure~\ref{fig:LightSimulation} shows some illustrative results for illuminants of different intensity and located at different relatives positions with respect to the face. 
\begin{algorithm}
\begin{algorithmic}[1]
\Procedure{addAmbientLight}{$I_0, z, p_0, l$} 
\Statex Make sure the depth map is smooth and noise-free. 
\State $z = $ Denoising$(z)$ \Comment{(Gradient op. amplifies noise.)}
\Statex Compute depth gradient.
\State $z_x, z_y =$ ComputeGradient$(z)$
\Statex Compute surface normals \eqref{eq:normal_vectors}.
\State $n = [-z_x, -z_y, 1]$
\State $n = \frac{n}{\|n\|_2}$
\Statex Additional light factor (due to the new source). 
\State $I_{+}[x,y] = p_0\, \max\left(n(x,y) \cdot l,\ 0\right)   $ 
\Statex For each color add the additional brightness.
\State $I[x,y,c] = I_0[x,y,c] (1 + I_{+}[x,y])$
\Statex Model camera saturation.
\State $I = $clip$(I, [0, 1])$
\State \textbf{return} $I$ 
\EndProcedure
\end{algorithmic}
\caption{Steps for the simulation of additional ambient light. The inputs represent: $I_0$ the facial albedo, $z$ the facial depth map, $p_0$ the intensity of the source light, $l$ the direction in which the light source is located. Examples are shown in Figure~\ref{fig:LightSimulation}.}\label{alg:diff_illumination}
\end{algorithm}
%

\section{Hardware implementation: a feasibility study}
\label{app:hardware implementation}
We tested a potential hardware implementation of the proposed ideas using commercially available hardware. To project the fringe pattern we used a projector EPSON 3LCD with native resolution $1920\times1080$, and a webcam Logitech C615 with native resolution $1280\times720$. The features of this particular hardware setup are independent of the proposed ideas. One could choose, for example, projecting infrared light, or design a specific setup to meet specific deployment conditions.

\begin{figure}[htp]
    \centering
    \includegraphics[width=.6\columnwidth]{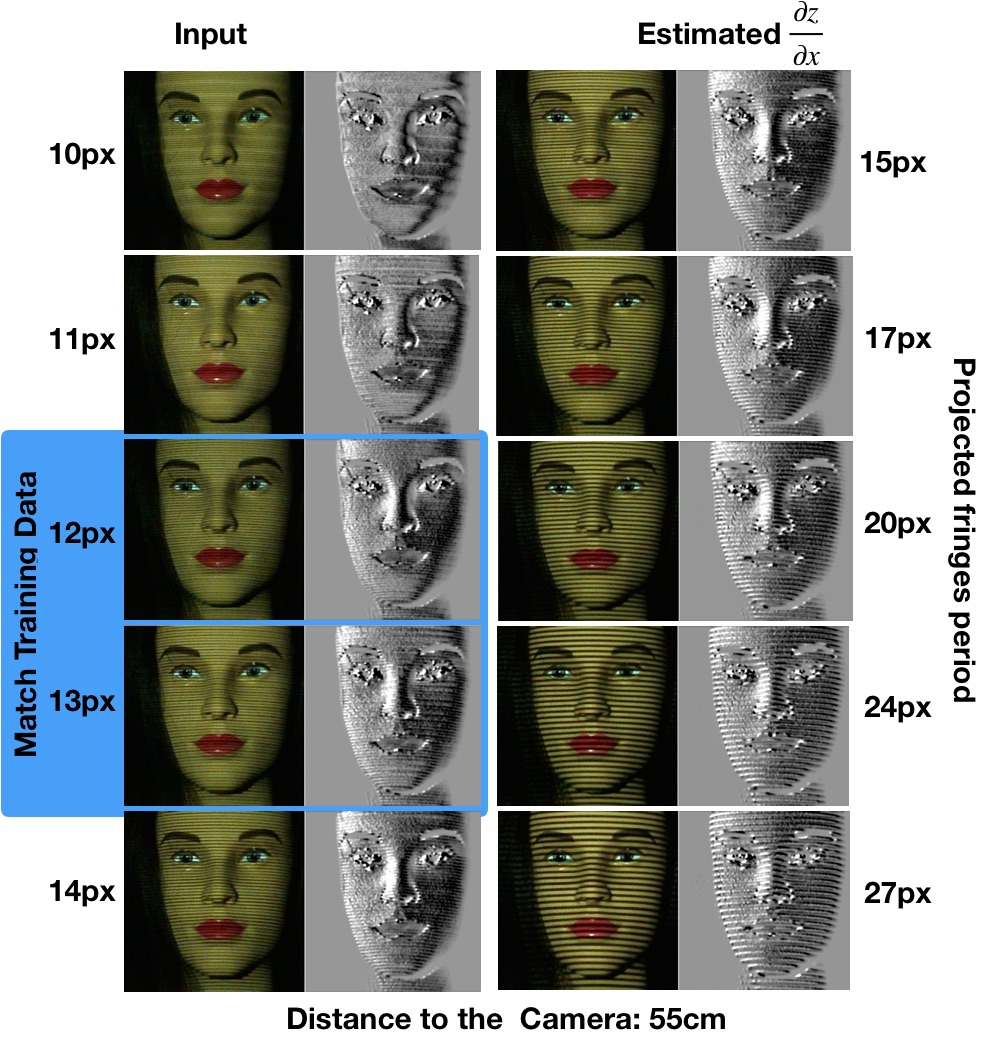}
    \caption{Testing the model on a hardware implementation. We projected the proposed light pattern using a commercial projector (EPSON 3LCD) and captured facial images using a standard webcam (Logitech C615). Side by side, we show the captured image (left) and the x-partial derivative of the depth estimated by our DNN model. We tested projected patters of different period, ranging from 10px to 27px (pixels measured in the projector sensor). }\label{fig:app_testing_fw}
\end{figure}

\begin{figure}[htp]
    \centering
    \includegraphics[width=.6\columnwidth]{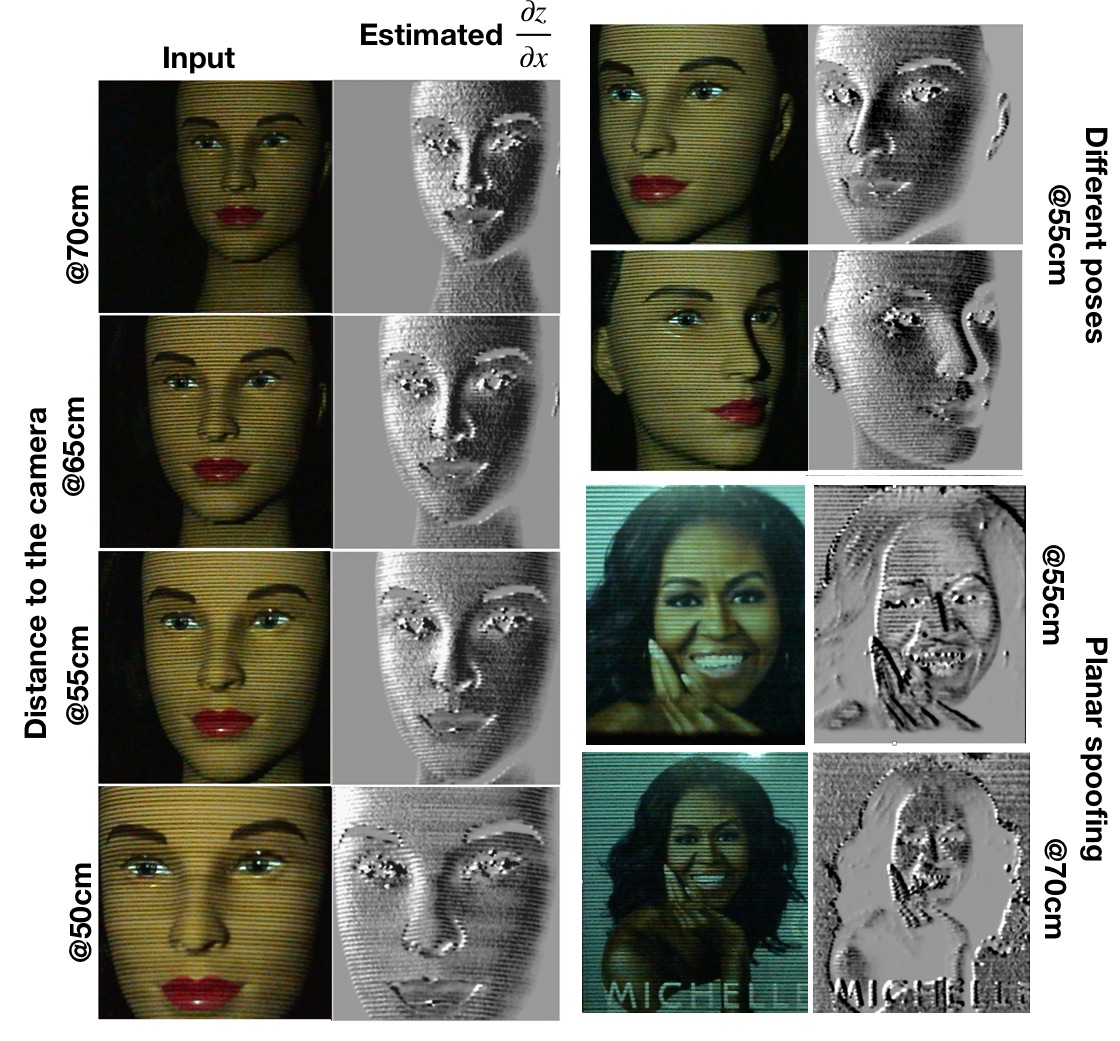}
    \caption{Changing the distance to the camera and pose. Complementing the results shown in Figure~\ref{fig:app_testing_fw}, we tested how our DNN model perform as the distance to the camera-project system changes. The left column shows side by side the image captured by the camera and the partial derivative of the depth estimated by our pre-trained model. The distance of the test face to the camera ranged from 50cm to 70cm. The left column shows on the top the results obtained for different head-poses, and on the bottom, the output when a planar facial portrait is presented.} \label{fig:app_testing_distance}
\end{figure}

Figure~\ref{fig:app_testing_fw} shows some empirical results obtained by photographing a mannequin at different relative positions and under different illumination. We qualitative tested how the width of the projected pattern affects the estimation performance. In addition, we tested (see Figure~\ref{fig:app_testing_distance}) how the distance of the face to the camera-projector system affects the prediction of the depth derivative. These two experiments are related as changing the distance to the camera changes the camera perception of the fringes width. 

It is important to take into account that in this experiments we tested our pre-trained models (completely agnostic to this particular hardware implementation), i.e., no fine-tuning or calibration was performed prior capturing the images presented in Figures~\ref{fig:app_testing_fw} and~\ref{fig:app_testing_distance}. In production settings, in contrast, one would fix a specific hardware setup, collect new ground truth data, and fine-tune the models to optimize the setup at hand. 